\documentclass[preprint,review,12pt]{elsarticle}

\usepackage{amssymb}
\usepackage{amsmath}

\usepackage{graphicx}
\usepackage[caption=false,font=footnotesize,labelfont=sf,textfont=sf]{subfig}
\usepackage{xcolor, soul,framed} 
\PassOptionsToPackage{hyphens}{url}\usepackage{hyperref}
\colorlet{shadecolor}{yellow}
\usepackage{array}
\usepackage{mdwmath}
\usepackage{mdwtab}
\usepackage{eqparbox}
\usepackage{url}
\usepackage{algorithm}
\usepackage{algorithmic}
\usepackage{multirow}  
\usepackage{booktabs} 
\usepackage{threeparttable}
\usepackage{bbding}
\usepackage{ulem}
\usepackage{float}
\usepackage{subfig}
\usepackage{makecell}
\makeatletter
\newcommand*\bigcdot{\mathpalette\bigcdot@{.5}}
\newcommand*\bigcdot@[2]{\mathbin{\vcenter{\hbox{\scalebox{#2}{$\m@th#1\bullet$}}}}}
\makeatother
\definecolor{newcolor}{rgb}{.8,.349,.1}
\usepackage{ragged2e}

\usepackage{graphicx}
\usepackage{textcomp}
\usepackage{bbm}
\usepackage{bm}
\newcommand{\etal}{\emph{et al. }}
\definecolor{FigureColor}{RGB}{0,138,218}
\newcommand{\fig}[1]{\textcolor{FigureColor}{Fig.#1}}
\definecolor{EqColor}{RGB}{0,138,218}
\newcommand{\eqn}[1]{\textcolor{EqColor}{Eq.{#1}}}
\definecolor{TabColor}{RGB}{0,138,218}
\newcommand{\tab}[1]{\textcolor{TabColor}{Table {#1}}}
\definecolor{Equation}{RGB}{0,138,218}
\definecolor{Table}{RGB}{0,138,218}
\definecolor{Algorithm}{RGB}{0,138,218}
\definecolor{Annotation}{RGB}{19, 139, 44}
\newcommand{\red}[1]{\textcolor{red}{#1}}
\newcommand{\green}[1]{\textcolor{green}{#1}}

\usepackage{bbding}

\usepackage{breqn}

\usepackage{wrapfig} %
\usepackage{flushend} %

\begin{document}

\begin{frontmatter}

\title{ReshapeIT: Reliable Shape Interaction with Implicit Template for Anatomical Structure Reconstruction}

\author[1]{Minghui Zhang\fnref{co-first}} 
\address[1]{Institute of Medical Robotics, Shanghai Jiao Tong University, Shanghai, 200240, China.}
\author[2]{Hao Zheng\fnref{co-first}} 
\address[2]{Jarvis Research Center, Tencent Youtu Lab, Shenzhen, China.}
\author[2]{Yawen Huang} 
\author[3]{Ling Shao} 
\address[3]{UCAS-Terminus AI Lab, University of Chinese Academy of Sciences, Beijing, China.}
\author[1]{Yun Gu\corref{cor}} 
\fntext[co-first]{These authors contributed equally to this work.}
\cortext[cor]{Corresponding author.}

\begin{abstract}
Shape modeling of volumetric medical images is crucial for quantitative analysis and surgical planning in computer-aided diagnosis. To alleviate the burden of expert clinicians, reconstructed shapes are typically obtained from deep learning models, 
such as Convolutional Neural Networks (CNNs) or transformer-based architectures, followed by the marching cube algorithm. However, automatic shape reconstruction often 
falls short of perfection due to the limited resolution of images and the absence of shape prior constraints. 
To overcome these limitations, we propose the Reliable Shape Interaction with Implicit Template (ReShapeIT) network, which models anatomical structures in continuous space rather than discrete voxel grids. 
ReShapeIT represents an anatomical structure with an implicit template field shared within the same category, complemented by a deformation field. 
It ensures the implicit template field generates valid templates by strengthening the constraint of the correspondence between the instance shape and the template shape. The valid template shape can then be utilized for implicit generalization. A Template Interaction Module (TIM) is introduced to reconstruct unseen shapes 
by interacting the valid template shapes with the instance-wise latent codes. Experimental results on three datasets demonstrate the superiority of our approach in anatomical structure reconstruction. 
The Chamfer Distance/Earth Mover's Distance achieved by ReShapeIT are 0.225/0.318 on Liver, 0.125/0.067 on Pancreas, and 0.414/0.098 on Lung Lobe. The source code is available at: \url{https://github.com/EndoluminalSurgicalVision-IMR/ISMM}.
\end{abstract}




\begin{keyword}
Anatomy Prior \sep Correspondence Constraint \sep Medical Shapes



\end{keyword}

\end{frontmatter}



\section{Introduction}\label{sec:intro}
Shape modeling of volumetric medical images is critical to computer-aided diagnosis for medical image analysis. The reconstructed shapes in three dimensions (3D) provide a spatial understanding of the organs or lesions, which are beneficial to clinical decision-making as well as for surgery planning and guidance ~\citep{krass2022computer,preuss2022using}. Furthermore, an advanced way to utilize the reconstructed shapes is through virtual reality head-mounted wearables.
\begin{figure}[t]
\centering
\includegraphics[width=0.5\linewidth]{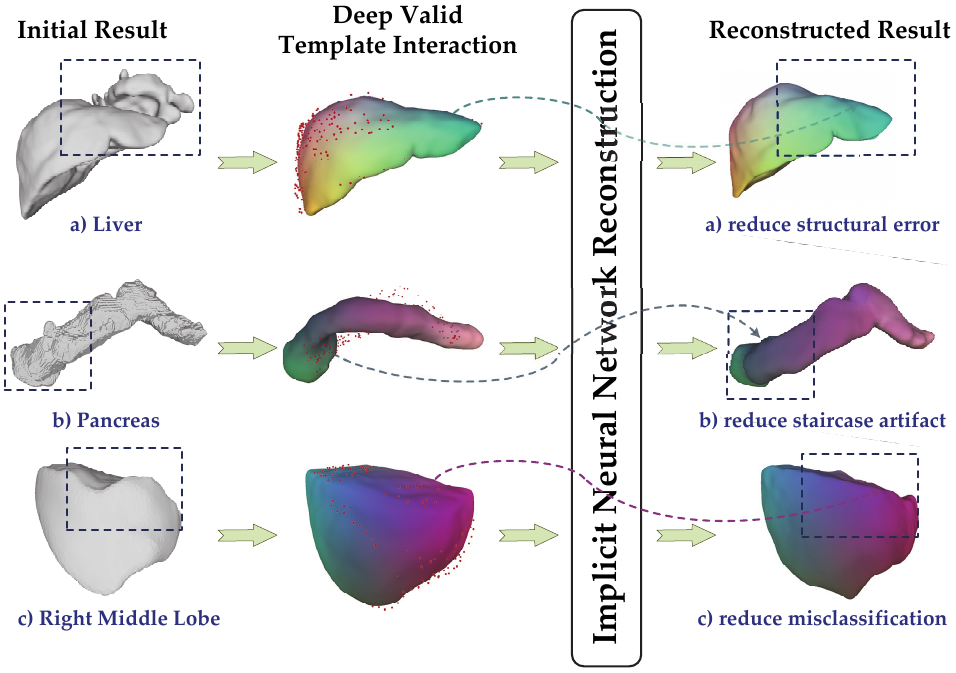}
\caption{ReShapeIT interacts initial results with valid template shapes based on reliable dense correspondence, followed by the implicit neural representation to acquire refined reconstruction.}
\label{fig:schematic_diagram}
\end{figure}
This allows users to deeply understand virtual 3D medical models and has been implemented in many clinical tasks and training scenarios~\citep{huber2017new, willaert2012recent}. 
However, it is extremely time-consuming for clinicians to delineate anatomical structures slice-by-slice. 
To relieve the burden of clinicians, the deep learning models, especially encoder-decoder structures~\citep{cciccek20163d,hatamizadeh2022unetr}, have shown great ability to automatically segment anatomical structure 
of medical images. Then, the anatomical shapes can be acquired by a posterior reconstruction via marching cube~\citep{lorensen1998marching} algorithm. 
The UNet network~\citep{cciccek20163d} are designed on top of CNNs with skip connection between encoder and decoder part.
The transformer-based architecture~\citep{hatamizadeh2022unetr} utilize a transformer module as the encoder instead of CNNs.
Although these structures exhibits great inductive bias ability, they encountered several challenges in medical shape modeling. 
First, CNNs or transformers may generate structural errors due to the lack of shape prior constraint.
Second, they are susceptible to image resolution. They can only approximate organ contours since they are limited to the discrete voxel grid. 
Third, due to the limit of GPU memory, current deep learning models adopts the patch-wise training and testing paradigm, which may fail to model the long-range dependency of anatomical structures. 
The aforementioned challenges lead to the structural errors (as seen in \fig{\ref{fig:schematic_diagram}.a)}, initial segmentation result of liver structure generated by 3D UNet), 
staircase artifact of the boundary (as seen in \fig{\ref{fig:schematic_diagram}.b)} initial result of pancreas), and misclassification problems in the shape modeling (as seen in \fig{\ref{fig:schematic_diagram}.c)} initial result of the right middle lobe).
Previous works designed consecutive procedure, e.g., cascaded UNet structures~\citep{jha2020doubleu}, adversarial structure correction~\citep{dai2018scan}, and variational auto encoder that improve 
topology coherence~\citep{araujo2019deep}, to refine the initial prediction results by CNNs. However, these methods still focus on the discrete voxel grid and have not exert strong shape constraint, hence, the improvement is limited. 
To alleviate the aforementioned challenges, we propose to utilize the implicit neural representation networks (INRs) to model the medical anatomical structures. 
An inherent advantage of INRs lies in their migration from the shape modeling within the discrete voxel grids to continuous physical space. There existed several works that attempted to leverage INRs for medical shape modeling ~\citep{marimont2022implicit,sorensen2022nudf,khan2022implicit,amiranashvili2024learning}. 
However, ~\cite{marimont2022implicit,sorensen2022nudf,khan2022implicit} merely added the coordinate-based latent constraint in the segmentation process, which does not particularly break the limitation of the discrete voxel grid. 
~\cite{amiranashvili2024learning} proposed the continuous shape modeling. However, the sharable structural prior has not been explored. Given that medical shapes share similarities across different instances, e.g., the liver structure can be divided into liver right lobe, liver left lobe, and the inferior hepatic margin, the pancreas shares the head, body, and tail parts.  
We hypothesize that the anatomical structure of an instance can be represented by a sharable implicit template field, together with its deformation field. Therefore, we propose the novel Reliable Shape Interaction
with Implicit Template (ReShapeIT) to model the medical anatomical structures in the continuous space, and simultaneously explore the valid template shape shared across the same category. 
Generally, the ReShapeIT is sequentially composed of an implicit deformation field and template field, and the signed distance function (SDF) is chosen as the constraint to facilitate the shape representation, similar to ~\cite{park2019deepsdf}. 
The world coordinates of the instance shapes, along with the assigned paired latent codes with each instance are first fed into the deformation field. The deformation field serves as the alignment function, 
generating the offset of the coordinates from the instance shape to the latent template shape, and the scalar displacement of SDF for each instance shape. Then the template field takes the coordinates with the offset as 
input to optimize the template shape. Based on the Deform-Template flow, the common structural prior knowledge shared across the same category of medical shapes can be explored. However, the generated template shape is not always valid because the 
constraint between the instance shape and the template shape is weak. To facilitate the validness of the template shape, we strengthen the constraint of the correspondence between the instance shape and the template shape. Specifically, 
ReShapeIT implicitly models the instances shapes, spanning the template space. Meanwhile, the template space can be sampled and assigned the shape latent code the same way as for the instances. Then the sampled template coordinates and 
template latent codes are also fed into the deformation field to reconstruct reliable template shapes, simultaneously, the dense correspondence between the instance and the template shape can be accurately established. 

Moreover, the reliable template shape facilitates the implicit generalization on unseen shapes. To take the advantage of the template shape, we propose the template interaction module (TIM) that interacts initial result from either CNNs or transformers 
with the reliable template shape. The points of the initial result with high confidence when aligning to the template shape are sampled. The coordinates of the chosen points are firstly input to the fine-tuned deformation field of the ReshapeIT to optimize the deformation field and latent codes, and then fed into the fixed template field to reconstruct the refined results.

To validate the effectiveness of the proposed method, we performed extensive experiments on three medical datasets, including liver, pancreas, and lung lobes. 
Compared with other state-of-the-art voxel-based / coordinate-based approaches, experimental results demonstrate that proposed method achieved superior quantitative performance with better interpretability. The main contributions of this work are summarized as follow:

1) The novel ReShapeIT is proposed for medical anatomical structure reconstruction in the continuous space. It breaks the limitation of the grid resolution of image scans and explores the sharable structural prior knowledge within the same category via the Deform-Template flow.

2) The correspondence between the reconstructed instance shape and the implicit template is strengthened to achieve reliable template shape.

3) Template Interaction Module (TIM) is proposed to generalize the reliable template shape to the reconstruction of unseen shapes.

\section{Related Works}\label{sec:related_works}
\subsection{Deep Implicit Surface Reconstruction}
Implicit functions ~\citep{carr2001reconstruction, shen2004interpolating} were adopted to represent shapes via constructing volumetric fields and expressing shapes of their iso-surfaces.
With the development of deep learning methods, deep implicit functions have been introduced into neural networks as neural fields ~\citep{park2019deepsdf,mescheder2019occupancy,sitzmann2020implicit,czerkawski2024neural,sitzmann2019scene,huang2024efficient} 
and have emerged as a powerful paradigm. DeepSDF ~\citep{park2019deepsdf} is a representative method that defines the surface of its shape as the level set of a signed distance field (SDF). 
Specifically, it defines a neural field, termed $\mathcal{F}_{\mathrm{SDF}}$. For a given spatial point $\bm{p} = \{x,y,z\} \in \mathbb{R}^{3}$, along with the specific latent code $\alpha \in \mathbb{R}^{\mathrm{K}}$ for each instance, 
$\mathcal{F}_{\mathrm{SDF}}$ takes them as input and output signed distance value, $\mathcal{F} _{\mathrm{SDF}}: \mathbb{R} ^{3+\mathrm{K}} \to \mathbb{R}$.
where the absolute value of $s$ denotes the distance to the closet surface, and the sign encodes whether $\bm{p}$ is inside (negative) and outside (positive) the shape surface. 
$\mathcal{F}_{\mathrm{SDF}}$ is a continuous implicit function since $\bm{p}$ is given arbitrarily of every possible 3D point rather than discrete 3D locations in voxel representation. 
With $\mathcal{F}_{\mathrm{SDF}}$, the shape surface can be implicitly represented based on the iso-surface of $\mathcal{F}_{\mathrm{SDF}}(\cdot) = 0$, followed by Marching Cube algorithm~\citep{lorensen1998marching} to extract the mesh.
In addition, some other approaches regarded the continuous implicit function as occupancy probability prediction~\citep{mescheder2019occupancy,peng2020convolutional,xu2019disn}. 
Recent studies take the consideration of local patterns in the complex shapes/scenes and introduce local implicit functions to capture geometric details~\citep{jiang2020local,chabra2020deep}.
However, current deep implicit functions focus on the shape reconstruction and completion task, and they need partial observation data even in unknown shape reconstruction tasks, which limits their 
generalization ability. Our proposed ReShapeIT generates valid template shapes and builds accurate dense correspondences among instances, which is beneficial to be generalized to the reconstruction on unseen instances.

\subsection{Latent Constraints for Shape Reconstruction}
Latent constraints are adopted as shape priors for the medical anatomical structure reconstruction.
Ricardo \etal~\citep{araujo2019deep} cascaded a variational autoencoder after the FCN to construct the  improved topology coherence network (ICTNet). It aims to learn a latent space that is capable of reducing topological incoherence. 
Jha \etal designed two consecutive U-Net structures to construct the Double-UNet~\citep{jha2020doubleu}. The output of the first UNet was regarded as the latent features, transported into the second UNet with the multiplication of the origin image. In addition, the features derived from the encoder of the first UNet were skip-connected to the decoder of the second UNet, followed by the two concatenated output feature maps obtained from the two corresponding UNets. 
Dai \etal proposed the Structure Correcting Adversarial Network (SCAN) ~\citep{dai2018scan} to refine organ segmentation. SCAN incorporates an adversarial network with the FCN to better preserve structural regularities inherent in human physiology. Similar to ~\cite{isola2017image}, the adversarial network maps the prediction and the ground-truth to the latent space, aiming to learn higher-order discriminative structures discriminate. Via the adversarial training process, the learned global information spreads backward to the FCN to achieve realistic outcomes. SCAN mainly reinforces the FCN to simulate the ground-truth while not designed to learn to correct structural errors based on the shape prior. 

Recent works attempted to utilize implicit representation in medical structure reconstruction~\citep{marimont2022implicit,sorensen2022nudf,khan2022implicit}. 
Implicit U-Net ~\citep{marimont2022implicit} adapted the implicit representation paradigm to volumetric medical images. They replaced the UNet Decoder branch with an implicit decoder proposed by ~\cite{park2019deepsdf}. It extracted features of a point $p$ from multiple spatial resolutions. The gather layer operates on each resolution of the encoder part to consistently acquire the same size of point-wise feature maps, which are then concatenated along with the original coordinates for the implicit decoder. 
Kristine \etal designed the NUDF~\citep{sorensen2022nudf}, aiming to represent the medical shape surface. Volumetric images are first processed through the encoder part of 3D-UNet to produce feature maps of multiple resolutions. Points are sampled from these feature maps, and then fed into a fully connected neural network to predict the distance from the point to the surface. 
Khan \etal designed the IOSNet ~\citep{khan2022implicit} that utilizes implicit neural representation for medical organ reconstruction. Compared with ~\cite{marimont2022implicit}, the point-wise feature maps are not recalibrated to the same size, and the implicit decoder part is more lightweight than the Implicit U-Net.
However, ~\cite{marimont2022implicit,sorensen2022nudf,khan2022implicit} merely added the coordinate-based latent constraint in the segmentation process, which does not particularly break the limitation of the discrete voxel grid. 
Stolt-Ans{\'o} \etal~ \citep{stolt2023nisf} and Amiranashvili \etal~\citep{amiranashvili2024learning} modeled the medical shapes in the continuous space in the training phase. 
In the test phase, They finetuned the latent shape codes for the reconstruction of unseen cases. However, the common structural prior knowledge shared across the same category of medical shapes has not been explored.

\begin{figure*}[t]
    \centering
    \includegraphics[width=0.90\linewidth]{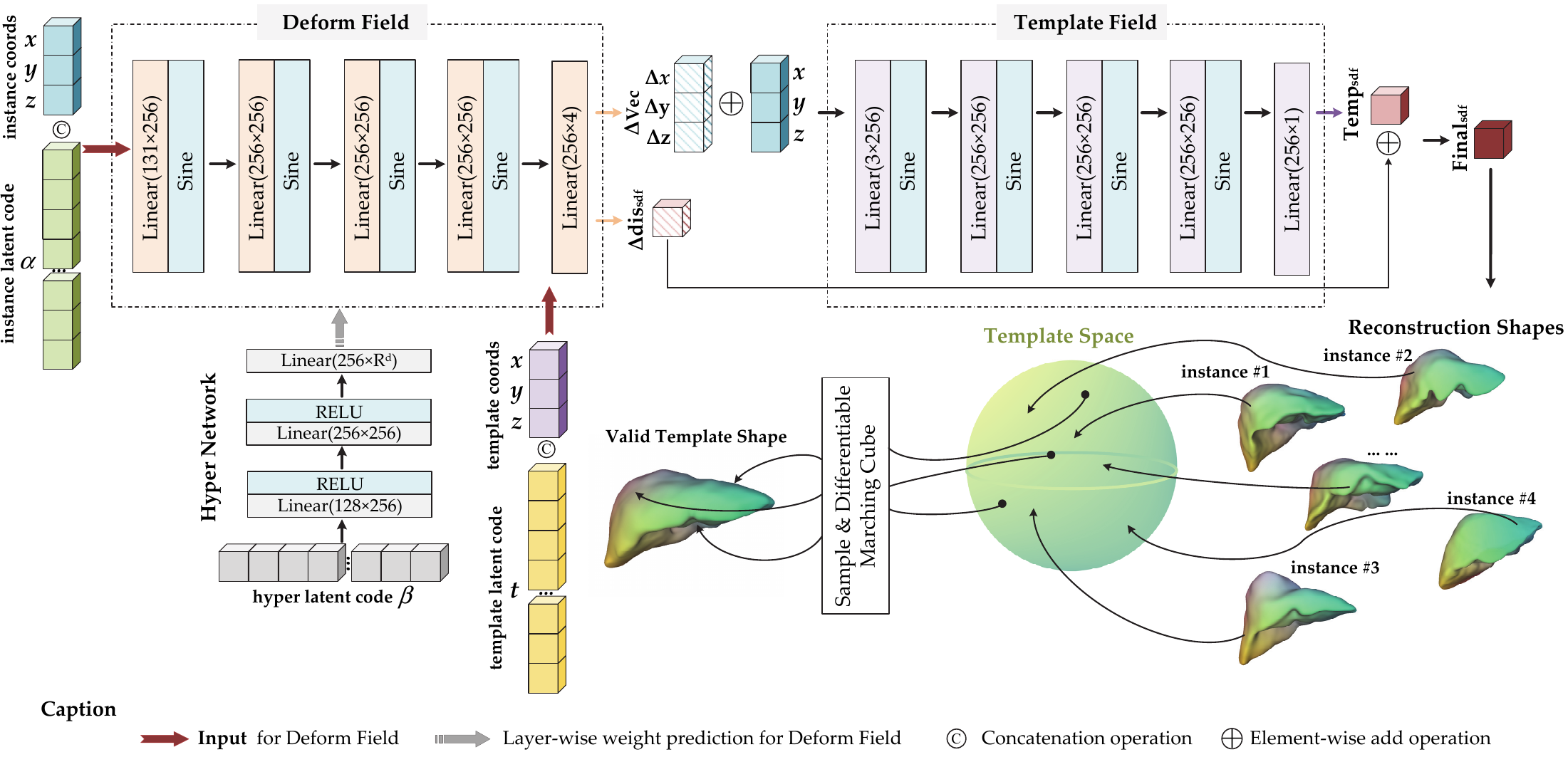}
    \caption{
    Overview of the Reliable Shape Interaction with Implicit Template (ReShapeIT) Framework. 
    For the instance shapes, world coordinates $\bm{p} = \{x,y,z\}$ along with corresponding latent code $\bm{\alpha}$ are fed into the implicit deform field $\mathit{Deform}$ 
    followed by one implicit template field $\mathit{Temp}$ to construct implicit shape modeling of medical anatomical structures. 
    The template space is sampled and assigned the latent template code $\bm{t}$, and they are also fed into the $\mathit{Deform}$ as the same way for instance shapes, aiming to build accurate correspondence between the template shape and instance shape, simultaneously achieve the reliable template shape.
    }
    \label{fig:framework}
    \end{figure*}
    \section{Methodology}\label{sec:method}
    \subsection{Implicit Deform-Template Shape Modeling}
    To address the limitation of the grid resolution of images and the lack of shape prior constraints, we propose the Reliable Shape Interaction with Implicit Template (ReShapeIT) network, which models anatomical structures in continuous space of the world coordinate system rather than discrete voxel grids. 
    It is well-acknowledged that the same category of medical shapes shares similar anatomical structures~~\citep{wang2012multi,iglesias2015multi}. 
    For example, the liver structure can be divided into liver right lobe, liver left lobe, and the inferior hepatic margin. The pancreas shares 
    the head, body, and tail parts. Inspired by recent INRs ~\citep{park2019deepsdf,mescheder2019occupancy,sitzmann2020implicit}, we decompose the paradigm of Deepsdf~\citep{park2019deepsdf}, as seen in \eqn{\ref{eq:sdf}}, into one implicit deformation field 
    followed by one implicit template field (Deform-Template) to construct implicit shape modeling of medical anatomical structures. 
    As illustrated in \fig{\ref{fig:framework}}, each instance shape is assigned by a learnable n-dimensional latent vector $\bm{\alpha} \in \mathbb{R}^{\mathrm{n}}$ to encode the desired shape space.
    Conceptually, we map the instance latent vector $\bm{\alpha}$, concatenated with the shape coordinates $p = \{x,y,z\} \in \mathbb{R}^{3}$ to the Deform-Template implicit field. 
    The deformation of each case to the template field should be instance-specific. Therefore, the hypernetwork is designed to predict the parameters of the deformation field for each instance. 
    Similar to ~\citep{david2017}, we introduce an additional hyper latent code $\bm{\beta} \in \mathbb{R}^{\mathrm{n}}$ for each instance to form a hyper-network $\mathcal{H}(\bm{\beta})$.
    $\mathcal{H}(\bm{\beta})$ aims to generate the weights of the deformation field, which can be seen in \fig{\ref{fig:framework}}. The overall design of Deform-Template implicit function, $\mathcal{F}_{\mathrm{D-T}}$, can be represented as:
    \begin{align}
    \mathcal{F}_{\mathrm{D-T}}\;(\bm{p},\bm{\alpha},\bm{\beta}) = \mathit{Temp} \circ \mathit{Deform} (\bm{p},\bm{\alpha},\bm{\beta}) \to  s , \label{eq:Deform-Template}
    \end{align}
    where $\mathcal{F}_{\mathrm{D-T}}: (\bm{p},\bm{\alpha},\bm{\beta}) \in \mathbb{R}^{3 + \mathrm{n}} \times \mathbb{R}^{\mathrm{n}} \to s \in \mathbb{R}$. $\mathcal{F}_{\mathrm{D-T}}$ is a composite 
    function formed by two implicit networks, $\mathit{Deform}$ and $\mathit{Temp}$. The weights of $\mathit{Deform}$ are derived by the hyper-network $\mathcal{H}$: $\mathcal{H}: {\bm{\beta}} \in \mathbb{R}^{\mathrm{n}} \to \mathit{Deform}_{layer^i} \in \mathbb{R}^{\mathrm{d}}.$.
    Both $\mathit{Temp}$ and $\mathit{Deform}$ are implemented by Multilayer Perceptrons (MLPs) with Sine activation ~\citep{sitzmann2020implicit}. 
    $\mathcal{H}$ comprises a set of MLPs with ReLU activation, each responsible for the weights of a single fully-connected layer $\mathit{i}$ within $\mathit{Deform}$. We abbreviate 
    the parameterized $\mathit{Deform}$ as $\mathit{Deform}_{\mathcal{H}(\bm{\beta})}$ in the following description. 
    
    $\mathit{Deform}_{\mathcal{H}(\bm{\beta})}$ takes the instance latent vector $\bm{\alpha}$ and continuous coordinate $p$ as input and transformed them to the latent template space by predicting 
    the deformation flow vector field $\Delta \bm{vec}$ and scalar displacement field $\Delta dis$:
    \begingroup
    \small
    \begin{align}
    \mathit{Deform}_{\mathcal{H}(\bm{\beta})}: (\bm{p},\bm{\alpha})  \to (\Delta \bm{vec} \in \mathbb{R}^{3}, \Delta dis \in \mathbb{R}). \label{eq:Deform}
    \end{align}
    \endgroup
    $\Delta \bm{vec}  = \{\Delta x, \Delta y, \Delta z \}$ denotes the per-point spatial offset that deforms the instance to the template space. The transformed coordinate can be calculated by $\bm{p} + \Delta \bm{vec}$, which is subsequently fed into the 
    $\mathit{Temp}$ to acquire the SDF value in the template space: $\mathit{Temp}: (\bm{p} + \Delta \bm{vec}) \in \mathbb{R}^{3} \to \hat{s} \in \mathbb{R}$, 
    where the $\hat{s}$ stands for the mapped SDF value of the instance in the template space, i.e., $\mathrm{Temp_{sdf}}$. $\Delta dis$ is the per-point displacement to handle anatomical variations. 
    It is added into the $\hat{s}$ to obtain final instance-specific SDF value $s$, i.e., $\mathrm{Final_{sdf}}$ . Hence, \eqn{\ref{eq:Deform-Template}} can be rewritten as:
    \begingroup
    \small
    \begin{align}
    \begin{aligned}
    s &= \mathcal{F}_{\mathrm{D-T}}\;(\bm{p},\bm{\alpha},\bm{\beta}) = \mathit{Temp}(\bm{p} + \Delta \bm{vec}) + \Delta dis \\
    &= \mathit{Temp}(\bm{p},\mathit{Deform}_{\mathcal{H}(\bm{\beta})}^{\Delta  \bm{vec}}(\bm{p},\bm{\alpha})) + \mathit{Deform}_{\mathcal{H}(\bm{\beta})}^{\Delta dis}(\bm{p},\bm{\alpha}). \label{eq:Instance-SDF}
    \end{aligned}
    \end{align}
    \endgroup
    Given the training shapes, we apply the SDF regression combined with SDF function attributes as constraints to learn continuous SDFs of these shapes, $\mathcal{F}_{\mathrm{D-T}}\;(\bm{p},\bm{\alpha},\bm{\beta})$ is 
    the predicted SDF value and the objective function can be defined as follows:
    \begingroup
    \small
    \begin{align}
    \begin{aligned}
    \mathcal{L}_{sdf} =  &\sum_{i}(
    \omega_{s}\sum_{\bm{p} \in \Omega_{i} } \left | \mathcal{F}_{\mathrm{D-T}}(\bm{p},\bm{\alpha},\bm{\beta})-\mathbf{s}(\bm{p})  \right | + \\
    &\omega_{n}\sum_{\bm{p} \in S_{i}}(1 - \mathrm{S_{cos}}(\nabla \mathcal{F}_{\mathrm{D-T}}(\bm{p}, \bm{\alpha}, \bm{\beta}),\mathbf{n}(\bm{p}))) + \\
    &\omega_{Eik}\sum_{\bm{p} \in \Omega_{i} } \left | \left \| \nabla \mathcal{F}_{\mathrm{D-T}}(\bm{p},\bm{\alpha},\bm{\beta})\right \|_{2}  - 1   \right | + \\
    &\omega_{\varphi} \sum_{\bm{p} \in \Omega_{i} \setminus S_{i}} \phi (\mathcal{F}_{\mathrm{D-T}}(\bm{p},\bm{\alpha},\bm{\beta}))), \label{eq:Loss_SDF}
    \end{aligned}
    \end{align}
    \endgroup
    where $\mathbf{s}(\bm{p})$ and $\mathbf{n}(\bm{p})$ indicate the ground-truth of SDF and surface normal value, respectively. $\Omega_{i}$ denotes the 
    3D shape space of the $i^{th}$ instance, and $S_{i}$ is the corresponding shape surface. In fact, points will be sampled both on the surface and free space to calculate $\mathcal{L}_{sdf}$. 
    The first term in \eqn{\ref{eq:Loss_SDF}} directly supervises the SDF regression results. The second term illustrates the surface normal 
    constraint. $\nabla$ represents the spatial gradient of the neural field, and the gradient function of the SDF equals the surface normal. 
    The cosine similarity ($\mathrm{S_{cos}}$) between gradient function and surface normal ground-truth is adopted to supervise the normal consistency. 
    The third term enforces the amplitude of the SDF gradient function to be 1 as determined by the Eiknonal equation~\citep{osher2004level}. The last term penalizes off-surface points with SDF prediction to zero:
    $ \phi (s) = \mathrm{exp}(-\delta \cdot \|s\|), \delta \gg 1.$ $\omega_{s},\omega_{n},\omega_{Eik},\omega_{\varphi}$ are weighting terms used in $\mathcal{L}_{sdf}$.
    
    With $\mathcal{F}_{\mathrm{D-T}}$, the 3D medical anatomical shape is represented by an implicit template field that is shared across the same category. The shape variance is captured by the 
    implicit deform field, which includes the deformation flow vector field and scalar displacement field for instance-specific shape modeling. 
    
    \subsection{Valid Template Generation and Interaction}
    
    \begin{figure}[t]
    \centering
    \includegraphics[width=0.5\linewidth]{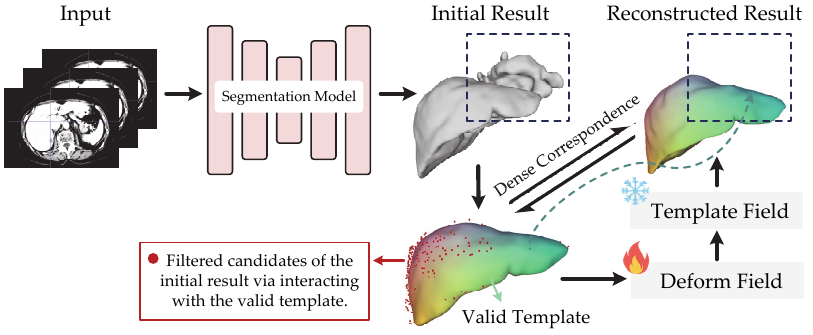}
    \caption{
    Template interaction module (TIM) for the implicit generalization for the reconstruction of unseen anatomical structures.}
    \label{fig:template_interaction_module}
    \end{figure}
        
    $\mathcal{F}_{\mathrm{D-T}}$ takes the advantage of continuous-wise modeling attribute of INRs, breaking the limitation of the image resolution and learning the shape prior knowledge from anatomical structure. 
    However, the learned shape prior, i.e., the template field is not always valid due to the weak constraint between the instance and the implicit template space. Therefore, we propose the valid template generation to 
    strengthen the correctness of the correspondence between the instances and the template field, aiming to acquire reliable template for the implicit generalization. Specifically, The reconstructed shapes 
    obtained from the implicit shape modeling can construct the template space, as demonstrated in \fig{\ref{fig:framework}}. Then the template space can be sampled points and converted to the plausible shape via differentiable 
    Marching Cube. We perform the point coordinate sampling on the template space $\bm{p_{t}} = \{x,y,z\}$, assign the template latent code $\bm{t}$ and hyper latent code $\bm{\beta_{t}}$ in the same way as for instances. 
    To strengthen the validness of the learned template, we feed the template coordinates and latent codes to the $\mathit{Deform}$: $\mathit{Deform}_{\mathcal{H}(\bm{\beta_{t}})}(\bm{p_{t}},\bm{t}) \to  \Delta TempShape = (\Delta \bm{vec_{temp}}, \Delta dis_{temp})$.
    We constrain both the $\Delta \bm{vec_{temp}}$ and $\Delta dis_{temp}$ to be zero. This latent template constraint guarantees the correctness of the dense correspondence between the template and instances, which improve the validness of the learned template.
    Further, the learned reliable template is beneficial to the implicit generalization. To generalize to the unseen shapes, we 
    propose the Template Interaction Module (TIM) to integrate learned reliable template shape prior into the initial result generated by CNNs. As illustrated in \fig{\ref{fig:template_interaction_module}}, 
    The valid template first interacts with the initial result to select the Top $\mathrm{K}\%$ matching candidate points based on the Coherent Point Drift algorithm ~\citep{myronenko2010point}. Then the filtered 
    points, along with the instance-wise latent codes are fed into the pretrained Deform-Template implicit network. The template field is frozen in the finetune stage while the deform field along with the instance-wise latent codes are 
    finetuned to optimize the unseen shapes. Benefited from the valid template and dense correspondence built by the implicit network, the reconstruction shape can be refined.
    
    
    \subsection{Model Optimization Pipeline}
    During the training stage, not only $\mathcal{L}_{sdf}$ is optimized to learn continuous SDF functions, but also additional constraints are introduced to facilitate the implicit shape modeling with template generation. 
    First, to optimize the implicit template to share common structures of the specific category, we add the normal direction constraints of points in the template space to be consistent with the corresponding given shape:
    
    \begingroup
    \small
    \begin{align}
    \begin{aligned}
    \mathcal{L}_{tpn} = \omega_{n} \sum_{i} {\sum_{\bm{p} \in S_{i}}(1 - \mathrm{S_{cos}}(\nabla \mathit{Temp}(\bm{p} + \Delta \bm{vec}),\mathbf{n}(\bm{p})))}.\label{eq:Loss_TemplateNormal}
    \end{aligned}
    \end{align}
    \endgroup
    Further, to prohibit the large shape distortion caused by $\Delta \bm{vec}$ and large displacement $\Delta dis$, we design two regularization terms, $\mathcal{L}_{vec}$ and $\mathcal{L}_{dis}$ to constrain $\mathit{Deform}_{\mathcal{H}(\bm{\beta})}$.
    $\mathcal{L}_{vec}$ aims to achieve smooth deform flow between the specific instance and the latent template. We add Laplacian smooth constraint to the $\Delta \bm{vec}$:
    \begin{align}
    \begin{aligned}
    &\mathcal{L}_{vec} = \sum_{i}{\sum_{\bm{p} \in \Omega_{i}} \left \|  \frac{\partial^{2}\mathit{D}_{\mathcal{H}(\bm{\beta})}^{\Delta \bm{vec}}}{\partial x^{2}} 
    + \frac{\partial^{2}\mathit{D}_{\mathcal{H}(\bm{\beta})}^{\Delta \bm{vec}}}{\partial y^{2}} + \frac{\partial^{2}\mathit{D}_{\mathcal{H}(\bm{\beta})}^{\Delta \bm{vec}}}{\partial z^{2}} \right \|_{2}}, \label{eq:Loss_Vec}
    \end{aligned}
    \end{align}
    $\mathit{D}_{\mathcal{H}(\bm{\beta})}^{\Delta \bm{vec}}$ is the short notion for $\mathit{Deform}_{\mathcal{H}(\bm{\beta})}^{\Delta \bm{vec}}(\bm{p},\bm{\alpha})$ in \eqn{\ref{eq:Loss_Vec}}. 
    In addition, $\mathcal{L}_{dis}$ aims to reduce the amplitude of displacement $\Delta dis$, enhancing the accuracy of implicit template and correspondence: $\mathcal{L}_{dis} = \sum_{i}{\sum_{\bm{p} \in \Omega_{i}}  \left | \mathit{Deform}_{\mathcal{H}(\bm{\beta})}^{\Delta dis}(\bm{p},\bm{\alpha}) \right |}$.
    Moreover, we constrain the template field to generate valid template shapes and accurate correspondence between the template and instances: $\mathcal{L}_{TempShape} = \left | \mathit{Deform}_{\mathcal{H}(\bm{\beta_{t}})}(\bm{p_{t}},\bm{t}) \right |$.
    Similar to ~\cite{park2019deepsdf}, we also assign the regularization loss to the instance latent codes $\bm{\alpha}$, template latent code $\bm{t}$, and the hyper latent code $\bm{\beta}$ both for instances and the template: 
    $\mathcal{L}_{reg\_\alpha} = \sum_{i}{ \left \| \bm{\alpha}_{i} \right \|_{2}^{2}}$, $\mathcal{L}_{reg\_t} = \left \| \bm{t} \right \|_{2}^{2}$, $\mathcal{L}_{reg\_\beta} = \sum_{i}{ \left \| \bm{\beta}_{i} \right \|_{2}^{2}} + \left \| \bm{\beta}_{t} \right \|_{2}^{2}$.
    
    
    
    
    In summary, the whole optimization of the implicit modeling with the valid template generation can be defined as:
    
    \begingroup
    \small
    \begin{align}
    \begin{aligned}
    \mathop{\arg\min}\limits_{\{ \bm{\alpha}, \mathit{Deform}_{\mathcal{H}(\bm{\beta})}, \mathit{Temp} \}}    &   \mathcal{L}_{sdf} + \mathcal{L}_{tpn} + \omega_{vec} \mathcal{L}_{vec} + \omega_{dis} \mathcal{L}_{dis} \\
    & + \omega_{dis} \mathcal{L}_{TempShape}  + \omega_{\alpha} \mathcal{L}_{reg\_\alpha} \\
    &+ \omega_{\alpha} \mathcal{L}_{reg\_t} + \omega_{\beta} \mathcal{L}_{reg\_\beta}, \label{eq:Loss_Total_train}  
    \end{aligned}
    \end{align}
    \endgroup
    where $\omega_{vec},\omega_{dis},\omega_{\alpha},\omega_{\beta}$ are the weighting terms. 
    As for the implicit generalization, 
    the learned implicit template is fixed at this stage. To embed the test shape $j$ to the latent space, specific instance latent code and the deform field are finetuned to acquire reconstructed shapes:
    \begin{align}
    \begin{aligned}
    \mathop{\arg\min}\limits_{\{ \bm{\alpha_{j}}, \mathit{Deform}_{\mathcal{H}(\bm{\beta_{j}})}\}} \mathcal{L}_{sdf} + \omega_{\alpha} \mathcal{L}_{reg\_\alpha_{j}}  + \omega_{\beta} \mathcal{L}_{reg\_\beta_{j}}. \label{eq:Loss_Total_finetune}  
    \end{aligned}
    \end{align}

\section{Experiments and Results}\label{sec:exp}
\subsection{Data}
\noindent\textbf{Dataset:} We evaluated our method on three public datasets: \textbf{MSD Liver Data\footnote{\url{https://registry.opendata.aws/msd/}}}. The liver dataset from the Medical Segmentation Decathlon (MSD)~\citep{antonelli2022medical} was used in our experiments. 
It contains 131 CT images in total,  91 CT images for training, 13 CT images for validation, and 27 CT images for testing. 
The in-plane resolution ranges from 0.5 to 1.0 mm, and the slice thickness ranges from 0.45 to 6.0 mm. The pixel values were clipped to [-100, 400] HU, and 
normalized to [0, 255]. \textbf{NIH Pancreas-CT Data\footnote{\url{https://www.kaggle.com/datasets/tahsin/pancreasct-dataset}}}. The pancreas dataset~\citep{roth2015deeporgan} contains 80 abdominal contrast 
enhanced 3D CT scans. 56 CT images for training, 8 CT images for validation, and 16 CT images for testing. 
The in-plane resolution ranges from 0.66 to 0.98 mm, and the slice thickness varies from 1.5 to 2.5 mm. 
The pixel values were clamped to [-100, 240] HU, before rescaled to [0, 255]. \textbf{Lung Lobe Data\footnote{\url{https://github.com/deep-voxel/}}}. The lung lobe dataset~\citep{tang2019automatic} that contains five lobes annotations was used in our experiments. 
It contains 50 CT scans, 35 CT images for training, 5 CT images for validation, and 10 CT images for testing. 
The in-plane resolution ranges from 0.54 to 0.88 mm, and the slice thickness varies from 0.625 to 2.5 mm. 
The pixel values were clamped to [-1000, 600] HU, before rescaled to [0, 255].

\makeatletter
\def\hlinew#1{%
\noalign{\ifnum0=1}\fi\hrule \@height #1 \futurelet
\reserved@a\@xhline}
\makeatother
\begin{table*}[t]
\renewcommand\arraystretch{1.0}
\centering
\caption{Quantitative comparison of the \textbf{Liver, Pancreas, and Lung Lobe} datasets in the continuous metric space. CD was multiplied by $10^2$.}
\label{tab:continuous_wise_result_three_datasets}
\scalebox{0.5}{
\begin{threeparttable}
\begin{tabular}{lcccc}
\hlinew{1.25pt}
\noalign{\vskip 1.5pt}
\multicolumn{5}{c}{\normalsize{\textbf{Liver Dataset}}}\\[1.5pt]\hlinew{0.5pt}
\textbf{Method} & \textbf{CD (mean. $\pm$ std.) $\downarrow$} & \textbf{EMD (mean. $\pm$ std.) $\downarrow$} & \textbf{CD (median) $\downarrow$} & \textbf{EMD (median) $\downarrow$}  \\ \hline
Double-UNet ~\citep{jha2020doubleu} & 2.823 $\pm$ 1.779  & 0.184 $\pm$ 0.047 &  2.912  & 0.196  \\
SCAN ~\citep{dai2018scan} & 2.125 $\pm$ 1.319 & 0.166 $\pm$ 0.049  & 2.304 & 0.177  \\
ICTNet ~\citep{araujo2019deep} & 0.504 $\pm$ 1.233 & 0.146 $\pm$ 0.037 & 0.802 & 0.123  \\
Implicit U-Net ~\citep{marimont2022implicit} & 0.536 $\pm$ 0.396 & 0.119 $\pm$ 0.018 & 0.512 & 0.110 \\
NUDF ~\citep{sorensen2022nudf} & 0.579 $\pm$ 0.774 & 0.118 $\pm$ 0.030 & 0.122 & 0.103 \\
IOSNet ~\citep{khan2022implicit}  & 1.820 $\pm$ 1.542 & 0.160 $\pm$ 0.048 & 1.096 & 0.157 \\ 
CSR \citep{amiranashvili2024learning}  & 0.382 $\pm$ 0.779  &  0.114 $\pm$ 0.045 &  0.125 & 0.100 \\ \hlinew{0.50pt}
\textbf{Proposed} & \textbf{0.225 $\pm$ 0.318} & \textbf{0.085 $\pm$ 0.017}& \textbf{0.083} & \textbf{0.078} \\
\hlinew{1.25pt}
\noalign{\vskip 1.5pt}
\multicolumn{5}{c}{\normalsize{\textbf{Pancreas Dataset}}}\\[1.5pt]\hlinew{0.5pt}
\textbf{Method} & \textbf{CD (mean. $\pm$ std.) $\downarrow$} & \textbf{EMD (mean. $\pm$ std.) $\downarrow$} & \textbf{CD (median) $\downarrow$} & \textbf{EMD (median) $\downarrow$}  \\ \hline
Double-UNet ~\citep{jha2020doubleu} & 1.084 $\pm$ 0.227  & 0.114 $\pm$ 0.067  & 0.154  & 0.089 \\
SCAN ~\citep{dai2018scan} & 0.240 $\pm$ 0.217 & 0.087 $\pm$ 0.018  & 0.157  & 0.081   \\
ICTNet ~\citep{araujo2019deep} & 0.578 $\pm$ 1.246  & 0.100 $\pm$ 0.045  & 0.191  &  0.086  \\
Implicit U-Net ~\citep{marimont2022implicit} & 2.190 $\pm$ 0.317  & 0.144 $\pm$ 0.083  &0.906  & 0.120 \\
NUDF ~\citep{sorensen2022nudf} & 0.310 $\pm$ 0.282  & 0.093 $\pm$ 0.024 & 0.237  & 0.091  \\
IOSNet ~\citep{khan2022implicit}  & 1.084 $\pm$ 2.554 & 0.106 $\pm$ 0.068  & 0.334 & 0.090 \\  
CSR \citep{amiranashvili2024learning}  & 0.218 $\pm$ 0.207 & 0.088  $\pm$ 0.028   & 0.146 & 0.081  \\ \hlinew{0.50pt}
\textbf{Proposed} & \textbf{0.125 $\pm$ 0.104} & \textbf{0.067 $\pm$ 0.024} & \textbf{0.098}  & \textbf{0.058}  \\
\hlinew{1.25pt}
\noalign{\vskip 1.5pt}
\multicolumn{5}{c}{\normalsize{\textbf{Lung Lobe Dataset}}}\\[1.5pt]\hlinew{0.5pt}
\textbf{Method} & \textbf{CD (mean. $\pm$ std.) $\downarrow$} & \textbf{EMD (mean. $\pm$ std.) $\downarrow$} & \textbf{CD (median) $\downarrow$} & \textbf{EMD (median) $\downarrow$}  \\ \hline
Double-UNet ~\citep{jha2020doubleu} & 0.792 $\pm$ 1.083 &  0.118 $\pm$ 0.036 & 0.324 & 0.107 \\
SCAN ~\citep{dai2018scan} & 0.772 $\pm$ 1.020  & 0.120 $\pm$ 0.034  & 0.511  & 0.111  \\
ICTNet ~\citep{araujo2019deep} &  0.746 $\pm$ 1.119   & 0.120 $\pm$ 0.036    & 0.426   &  0.109   \\
Implicit U-Net ~\citep{marimont2022implicit} & 0.809 $\pm$ 0.943  & 0.122 $\pm$ 0.036 & 0.420 & 0.113 \\
NUDF ~\citep{sorensen2022nudf} &  0.737 $\pm$ 1.006  & 0.117 $\pm$ 0.037 & 0.285 & 0.106  \\
IOSNet ~\citep{khan2022implicit}  &  0.869 $\pm$ 0.822  & 0.127 $\pm$ 0.030  & 0.526 & 0.124 \\
CSR \citep{amiranashvili2024learning}  & 0.537 $\pm$ 0.842  & 0.114 $\pm$ 0.030  & 0.268 & 0.110   \\ \hlinew{0.50pt}
\textbf{Proposed} & \textbf{0.414 $\pm$ 0.626}   & \textbf{0.098 $\pm$ 0.032}   &  \textbf{0.160}   & \textbf{0.088}  \\
\hlinew{1.25pt}
\end{tabular}
\end{threeparttable}}
\end{table*}

\noindent\textbf{Data Preparation:} The data preparation for the ReShapeIT should sample appropriate points with corresponding SDF values. Since all medical shapes are annotated as 3D volumes, we first 
extracted the surface mesh via Marching Cube algorithm~\citep{lorensen1998marching}.We followed a similar process of \cite{park2019deepsdf} 
to randomly sample points on the surface mesh and in the free space. Concretely, we first normalized each surface mesh to the unit sphere. Next, for surface points sampling, 
100 virtual images were rendered for each normalized mesh from 100 virtual cameras. Surface points were acquired through back-projecting the depth pixels, and the corresponding normals 
were assigned from the triangle to which it belongs. As for free space points sampling, we sampled them in the unit sphere uniformly and calculated the distance to the nearest surface. 
The sign of the distance was decided based on the depth with regard to the surface points. As long as the free points had a smaller depth value in any virtual image, it was assigned 
a positive sign, otherwise, it got a negative sign. In summary, 800K surface points along with normals and 200K free space points with corresponding SDF values were sampled for each shape.

\begin{figure}[t]
\centering
\includegraphics[width=0.8\linewidth]{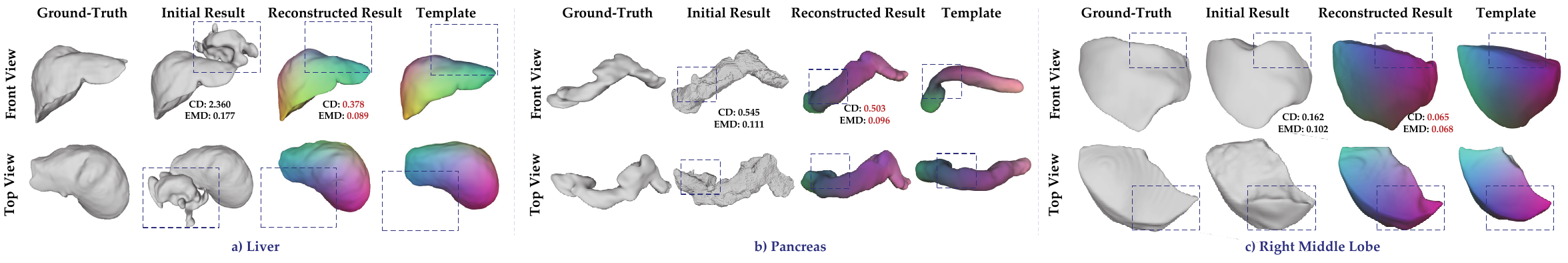}
\caption{
Examples of the implicit reconstruction results on the Liver, Pancreas, and Right Middle Lobe. The dotted blue boxes highlight the 
interaction areas between the initial result and the valid template shape. 
}
\label{fig:representative_reconstruction_refinement}
\end{figure}
\begin{figure}[t]
\centering
\includegraphics[width=0.80\linewidth]{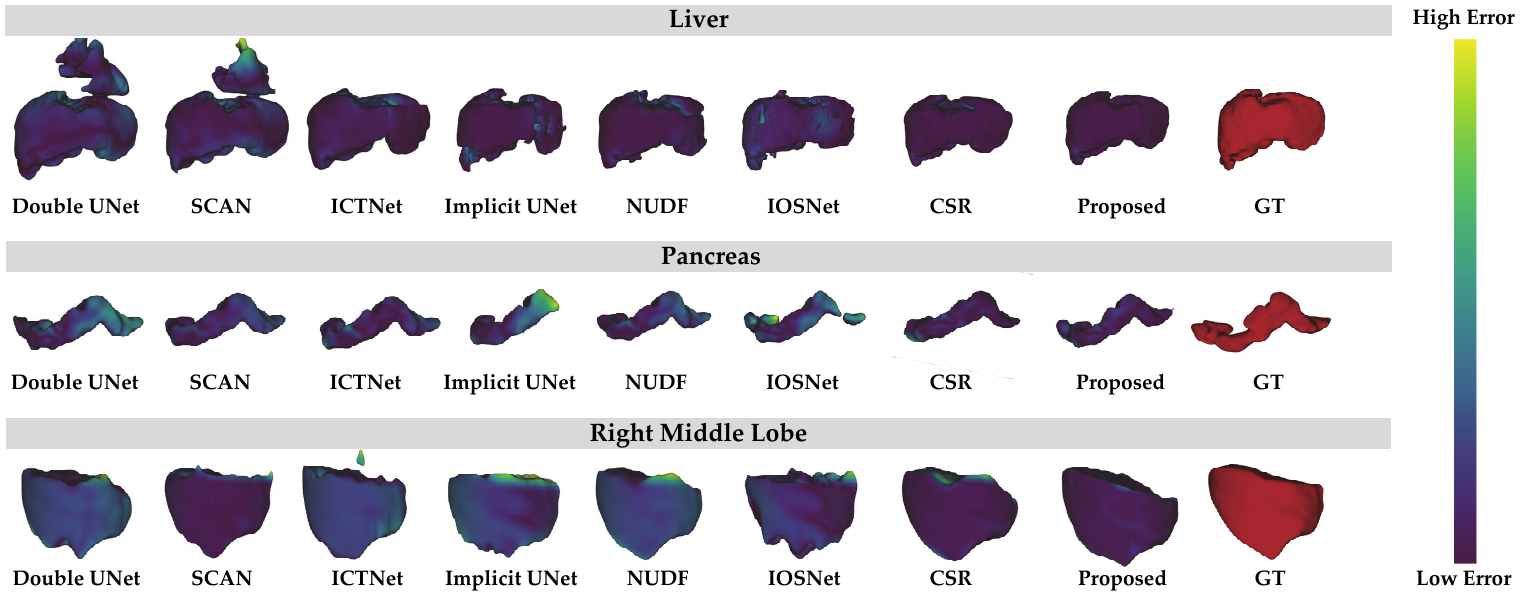}
\caption{Qualitative comparisons of the reconstruction error among the proposed methods with others. The dark purple indicates low reconstruction error, while bright yellow denotes high 
reconstruction error. }
\label{fig:reconstruction_error}
\end{figure}
\subsection{Implementation Details}
\subsubsection{Network Architecture Details}
Both the implicit neural network $\mathit{Temp}$ and $\mathit{Deform}$ were implemented by the MLPs. The dimension of instance latent code $\bm{\alpha}$, template latent code $\bm{t}$, and 
hyper latent code $\bm{\beta}$ were set to 128. The hyper-network $\mathcal{H}(\bm{\beta})$ consists of multiple 3-layer MLPs, each responsible for the weights of a 
single fully-connected layer $i$ within the implicit neural network $\mathit{Deform}$. The hidden feature is 256 in hyper-network $\mathcal{H}(\bm{\beta})$. 
Both $\mathit{Temp}$ and $\mathit{Deform}$ owns 5 fully-connected layers, of which the hidden feature was also set to 256. Sine activations were adopted in 
both $\mathit{Temp}$ and $\mathit{Deform}$ as proposed by \cite{sitzmann2020implicit}. Initialization of weights was also obeyed \cite{sitzmann2020implicit}. 
$\bm{\alpha}$ and $\bm{\beta}$ were initialized using $\mathcal{N}(0, 0.01^{2})$. ReLU activations were adopted in $\mathcal{H}(\bm{\beta})$ .

\subsubsection{Training and Finetune Details}
As for the training procedure, we jointly optimized all instance-wise latent codes $\bm{\alpha}$, $\mathcal{H}(\bm{\beta})$, and $\mathit{Temp}$ based on \eqn{\ref{eq:Loss_Total_train}}.  
The setting of weighting terms in $\mathcal{L}_{sdf}$ were followed by \cite{sitzmann2020implicit},  $\omega_{s}$, $\omega_{n}$, $\omega_{Eik}$, and $\omega_{\varphi}$ were 
set to $3e3$, $1e2$, $5e1$, and $5e2$ respectively. $\omega_{vec} = 5.0$, $\omega_{dis} = 1e2$, $\omega_{\alpha} = 1e5$, and $\omega_{\beta} = 1e6$ was used in the training progress. 
For each iteration, 4K surface points and 4K free-space points for each shape and the latent template space were randomly sampled for optimization. The Adam optimizer with a learning rate
of $1e-4$ and a batchsize of 16 were adopted. 200 epochs were trained in total. As for the finetune stage, since the implicit template field $\mathit{Temp}$ is shared across the class, 
we fixed its weight and adopted the learned template for interaction usage. The top $\mathrm{K}$ \% used in TIM module was set to Top 25 \%. $\bm{\alpha}$ and $\mathcal{H}(\bm{\beta})$ 
were finetuned via \eqn{\ref{eq:Loss_Total_finetune}} for 30 epochs. Other settings were identical to the training procedure. 
PyTorch framework was adopted to implement all experiments, executed on a Linux workstation with Intel Xeon Gold 5218 CPU @ 2.30 HZ, 128GB RAM, and 2 NVIDIA Geforce RTX 3090 GPUs.

\makeatletter
\def\hlinew#1{%
\noalign{\ifnum0=1}\fi\hrule \@height #1 \futurelet
\reserved@a\@xhline}
\makeatother
\begin{table*}[t]
\renewcommand\arraystretch{1.0}
\centering
\caption{Quantitative comparison of the \textbf{Liver, Pancreas, and Lung Lobe} datasets in the voxel-wise metric space.}
\label{tab:discrete_wise_result_three_datasets}
\scalebox{0.5}{
\begin{threeparttable}
\begin{tabular}{lcccc}
\hlinew{1.25pt}
\noalign{\vskip 1.5pt}
\multicolumn{5}{c}{\normalsize{\textbf{Liver Dataset}}}\\[1.5pt]\hlinew{0.5pt}
\textbf{Method} & \textbf{DSC (\%) $\uparrow$} & \textbf{NSD (\%) $\uparrow$} & \textbf{HD95 (mm) $\downarrow$} & \textbf{ASSD (mm) $\downarrow$}  \\ \hline
Double-UNet ~\citep{jha2020doubleu} & 82.40 $\pm$ 17.58  & 70.77 $\pm$ 16.83  & 81.06 $\pm$ 61.51  & 16.42 $\pm$ 38.47\\
SCAN ~\citep{dai2018scan}  & 86.68 $\pm$ 5.94 & 74.02 $\pm$ 10.23 & 60.05 $\pm$ 35.92 & 7.19 $\pm$ 4.23 \\
ICTNet ~\citep{araujo2019deep}  & 87.78 $\pm$ 6.45 & 71.61 $\pm$ 10.23 & 38.30 $\pm$ 28.50 & 5.54 $\pm$ 3.62  \\
Implicit U-Net ~\citep{marimont2022implicit} & 91.13 $\pm$ 5.09  & 74.03 $\pm$ 12.9  & 22.66 $\pm$ 10.58 & 3.25 $\pm$ 1.59 \\
NUDF ~\citep{sorensen2022nudf} & 93.39 $\pm$ 5.49  & 85.47 $\pm$ 12.0  & 17.20 $\pm$ 19.22 & 2.41 $\pm$ 2.14 \\
IOSNet ~\citep{khan2022implicit} & 86.48 $\pm$ 6.52  & 61.98 $\pm$ 14.67 & 49.11 $\pm$ 34.31 & 7.13 $\pm$ 4.70 \\ 
CSR \citep{amiranashvili2024learning}  & 93.24 $\pm$ 3.01   & 88.58 $\pm$ 8.83  & 15.57 $\pm$ 25.33  & 2.20 $\pm$ 1.54 \\ \hlinew{0.50pt}
\textbf{Proposed} & \textbf{93.91 $\pm$ 3.20} & \textbf{90.80 $\pm$ 7.61} & \textbf{10.92 $\pm$ 10.20} & \textbf{2.06 $\pm$ 1.28}\\
\hlinew{1.25pt}
\noalign{\vskip 1.5pt}
\multicolumn{5}{c}{\normalsize{\textbf{Pancreas Dataset}}}\\[1.5pt]\hlinew{0.5pt}
\textbf{Method} & \textbf{DSC (\%) $\uparrow$} & \textbf{NSD (\%) $\uparrow$} & \textbf{HD95 (mm) $\downarrow$} & \textbf{ASSD (mm) $\downarrow$}  \\ \hline
Double-UNet ~\citep{jha2020doubleu} & 79.40 $\pm$ 11.19  & 87.47 $\pm$ 12.64  &  13.23 $\pm$ 22.52 & 2.31 $\pm$ 3.70 \\
SCAN ~\citep{dai2018scan}  & 81.90 $\pm$ 6.29   & 90.24 $\pm$ 6.69   &  5.64 $\pm$ 3.57 & 1.20 $\pm$ 0.57  \\
ICTNet ~\citep{araujo2019deep}  & 79.54 $\pm$ 7.26   & 88.18 $\pm$ 9.21  &  7.95 $\pm$ 10.30   &  1.55 $\pm$ 1.32   \\
Implicit U-Net ~\citep{marimont2022implicit} & 68.47 $\pm$ 11.30  & 77.31 $\pm$ 13.28  & 19.08 $\pm$ 19.61  &  3.26 $\pm$ 3.46 \\
NUDF ~\citep{sorensen2022nudf} & 82.14 $\pm$ 5.17  & 90.46 $\pm$ 5.14 & 6.44 $\pm$ 5.73  & 1.26 $\pm$ 0.45  \\
IOSNet ~\citep{khan2022implicit} & 73.12 $\pm$ 13.72  & 84.40 $\pm$ 14.40 & 10.08 $\pm$ 17.00 & 2.27 $\pm$ 3.36 \\
CSR \citep{amiranashvili2024learning}  & 82.96  $\pm$ 5.84   & 91.84  $\pm$ 6.01 &4.81 $\pm$ 2.24 & 1.12 $\pm$ 0.43 \\ \hlinew{0.50pt}
\textbf{Proposed} & \textbf{83.44 $\pm$ 5.76}  & \textbf{92.45 $\pm$ 5.09}   & \textbf{4.08 $\pm$ 2.14}   &\textbf{1.05 $\pm$ 0.35}   \\
\hlinew{1.25pt}
\noalign{\vskip 1.5pt}
\multicolumn{5}{c}{\normalsize{\textbf{Lung Lobe Dataset}}}\\[1.5pt]\hlinew{0.5pt}
\textbf{Method} & \textbf{DSC (\%) $\uparrow$} & \textbf{NSD (\%) $\uparrow$} & \textbf{HD95 (mm) $\downarrow$} & \textbf{ASSD (mm) $\downarrow$}  \\ \hline
Double-UNet ~\citep{jha2020doubleu} &  92.55 $\pm$ 7.00  &  82.71 $\pm$ 12.65  &  15.97 $\pm$ 14.79 &  2.20 $\pm$ 2.04  \\
SCAN ~\citep{dai2018scan}  & 92.52 $\pm$ 7.00   &  83.53 $\pm$ 10.28   &  15.78 $\pm$ 12.68   &  2.10 $\pm$ 1.63    \\
ICTNet ~\citep{araujo2019deep}  & 92.59 $\pm$ 8.00   & 83.81 $\pm$ 11.29    & 16.36 $\pm$ 16.42  & 2.20 $\pm$ 2.08      \\
Implicit U-Net ~\citep{marimont2022implicit} & 89.59 $\pm$ 14.00  & 79.70 $\pm$ 16.39  & 18.43 $\pm$ 16.27  & 2.95 $\pm$ 3.44  \\
NUDF ~\citep{sorensen2022nudf} & 93.12 $\pm$ 7.00 &  84.82 $\pm$ 10.65 &  19.76 $\pm$ 19.23 & 2.38 $\pm$ 2.30 \\
IOSNet ~\citep{khan2022implicit} & 87.06 $\pm$ 16.00  & 73.43 $\pm$ 17.07  & 21.59 $\pm$ 15.17 & 3.93 $\pm$ 4.31 \\
CSR \citep{amiranashvili2024learning}  &  92.54 $\pm$ 4.59   &  85.22  $\pm$ 7.62   & 13.30 $\pm$ 10.58   &  2.05  $\pm$ 1.09 \\ \hlinew{0.50pt}
\textbf{Proposed} &  \textbf{93.53 $\pm$ 3.93}     &   \textbf{86.07 $\pm$ 5.68}      &   \textbf{11.75 $\pm$ 8.59}     &    \textbf{1.78 $\pm$ 0.92}   \\
\hlinew{1.25pt}
\end{tabular}
\end{threeparttable}}
\end{table*}

\subsubsection{Evaluation Metrics}
The principal metric of concern in this study is the quality of the shape surface that is refined by implicit neural networks in the continuous space. 
Chamfer Distance (CD) and Earth Mover's Distance (EMD) are chosen to measure the accuracy of the reconstruction. Detailed calculation can be found in the Supplementary Material.
In addition, the reconstructed shapes in the continuous space can be mapped back to the discrete voxel space. Hence, the secondary volume-based metrics are also included. We reported Dice Similarity Coefficient (DSC, \%),  Normalized Surface Dice~\citep{nikolov2018deep} (NSD, \%), 95\% Hausdorff distance (HD95, mm), and Average symmetric surface distance (ASSD, mm) in the quantitative analysis.

\begin{figure}[t]
\centering
\includegraphics[width=0.5\linewidth]{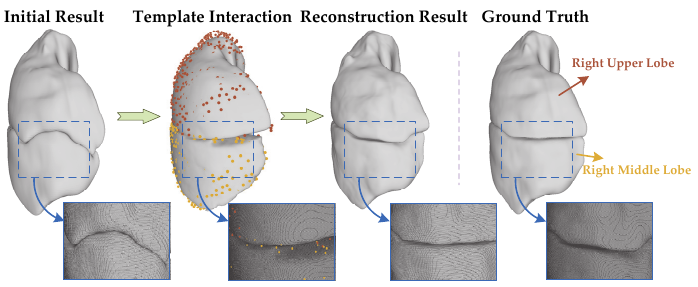}
\caption{Visualization of the multi lobes' reconstruction process via valid template interaction and implicit shape modeling. The example shows the interaction of the right upper lobe and the right middle lobe. 
the misclassification problem happens in the initial results. The template encodes each shape prior to the interaction, the reconstructed result demonstrates the improvement that they are closer to their independent ground-truth.}
\label{fig:multi_lobes_refinement}
\end{figure}
\begin{figure}[t]
\centering
\includegraphics[width=0.5\linewidth]{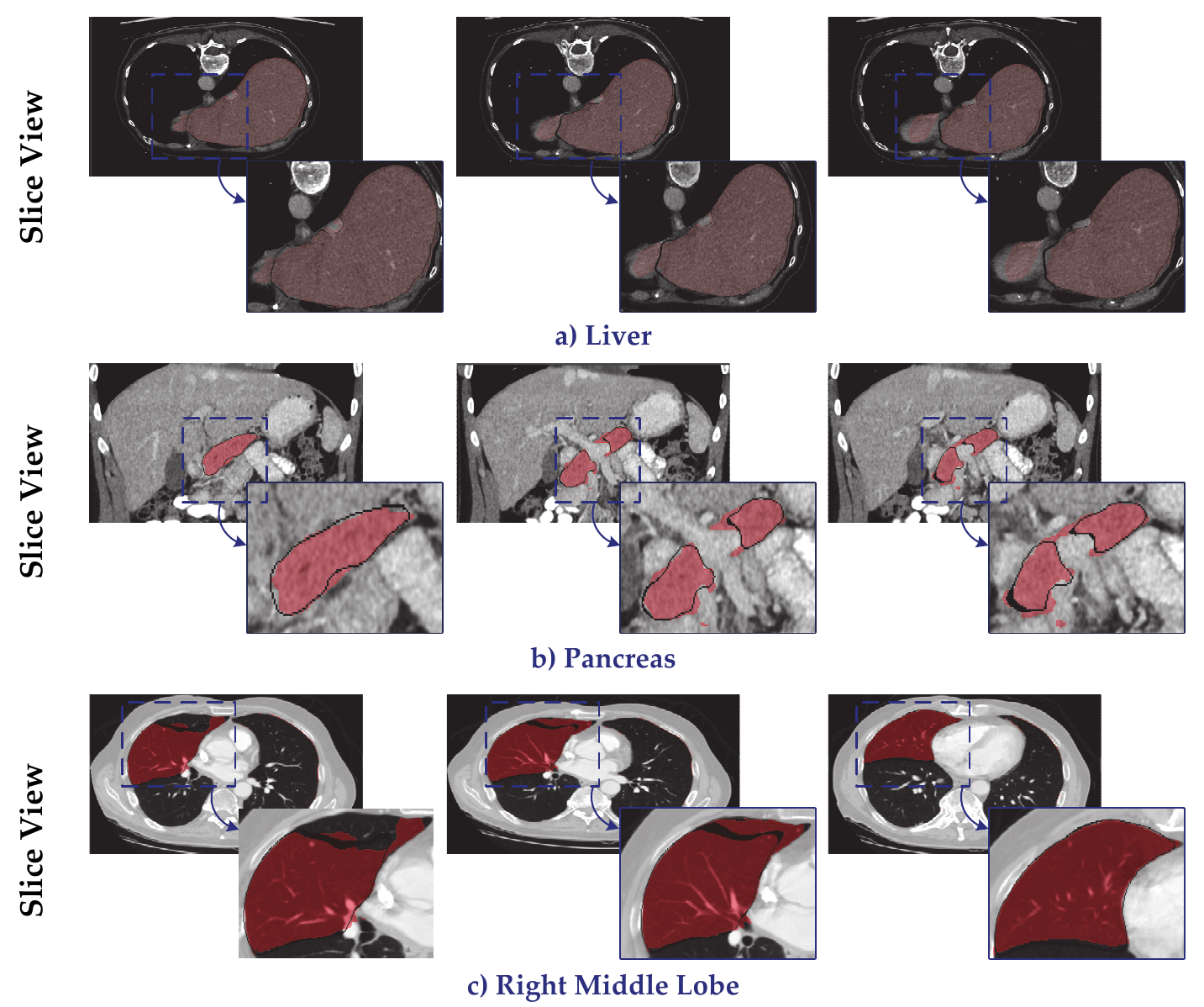}
\caption{Volume space comparison, results are shown in stacks of slices. The black contours denote the boundary results obtained from implicit reconstruction. Initial prediction results are labeled in red.}
\label{fig:slice_comparison}
\end{figure}


\subsection{Analysis of Unseen Shape Implicit Reconstruction}
The standard 3D UNet~\citep{cciccek20163d} was chosen as the segmentation model. Equipped with the implicit shape modeling, the results can be refined from an average 0.407 CD / 0.113 EMD to 0.225 CD / 0.085 EMD on the Liver dataset, from 0.223 CD / 0.097 EMD to 0.125 CD / 0.067 EMD on the Pancreas dataset, and from 0.677 CD / 0.115 EMD to 0.414 CD / 0.098 EMD on the Lung Lobe dataset. 
To further investigate the effectiveness of the proposed ReShapeIT, we provided visualization examples of the reconstruction results on the Liver, Pancreas, and Right Middle Lobe in \fig{\ref{fig:representative_reconstruction_refinement}}. 
The dotted blue boxes highlighted the refined areas, the same spatial location in the initial result, implicit reconstruction result, and template space. The CD and EMD were also decreased upon the implicit reconstruction. According to the learned template, the succeeding implicit reconstruction could alleviate the structural errors, staircase artifacts, and misclassification problem. 
In addition, recent latent constraint methods, including both the voxel-wise methods~\citep{araujo2019deep,jha2020doubleu,dai2018scan} 
and implicit-based methods ~\citep{marimont2022implicit,sorensen2022nudf,khan2022implicit,amiranashvili2024learning} were adopted for comparisons in our experiments.

\makeatletter
\def\hlinew#1{%
\noalign{\ifnum0=1}\fi\hrule \@height #1 \futurelet
\reserved@a\@xhline}
\makeatother
\begin{table*}[htbp]
\renewcommand\arraystretch{1.0}
\centering
\caption{Quantitative comparison of multi-class lobe dataset in continuous metric space and voxel-wise metric space.}\label{tab:multilobe_result}
\scalebox{0.4}{
\begin{threeparttable}
\begin{tabular}{@{}lcccccccccccccccccccc@{}}
\hlinew{1.25pt}
\noalign{\vskip 1.5pt}
\multicolumn{21}{c}{\normalsize{\textbf{Continuous Metric Space}}}\\[1.5pt]\hlinew{0.5pt}
\multirow{2}{*}{\textbf{Method}}& \multicolumn{5}{c}{\textbf{CD (mean. $\pm$ std.) $\downarrow$}} & \multicolumn{5}{c}{\textbf{EMD (mean. $\pm$ std.) $\downarrow$}} & \multicolumn{5}{c}{\textbf{CD (median) $\downarrow$}} & \multicolumn{5}{c}{\textbf{EMD (median) $\downarrow$}}  \\ \cmidrule(l){2-6} \cmidrule(l){7-11} \cmidrule(l){12-16} \cmidrule(l){17-21} 
                                                 &  RU   & RM   & RL   & LU   &  LL &  RU   & RM   & RL   & LU   &  LL    & RU   & RM   & RL   & LU   &  LL   & RU   & RM   & RL   & LU   &  LL  \\ \hline
Double-UNet ~\citep{jha2020doubleu}                &  0.660   & 1.320   & 0.848   & 0.286   & 0.844  & 0.117   & 0.142   & 0.126   & 0.093  & 0.114   & 0.346   & 1.265   & 0.262  & 0.078   &  0.192  & 0.107   & 0.145  & 0.115   &  0.087  & 0.096 \\
SCAN ~\citep{dai2018scan}                          &  0.605   &  1.361  & 0.774   & 0.272   & 0.846  &  0.119  & 0.142   & 0.127   & 0.097  & 0.113   &  0.563  & 1.024   & 0.382  & 0.100    &  0.198  &  0.111  & 0.136   & 0.118   & 0.092 & 0.098 \\
ICTNet ~\citep{araujo2019deep}                     &  0.498  & 1.467  &  0.600  & 0.249   & 0.917  & 0.114   &  0.150  &  0.124  & 0.095  & 0.115   &  0.438  & 1.539   & 0.276  & 0.119   & 0.139   & 0.106   &  0.156  &  0.115  & 0.098   & 0.106 \\
Implicit U-Net ~\citep{marimont2022implicit}       &  0.686   &  1.367  & 0.707   &  0.403  & 0.081  & 0.125   & 0.142   & 0.125   & 0.096  &  0.100  &  0.410  &  1.064  & 0.226  & 0.107   & 0.186   &  0.115  & 0.135    & 0.106   & 0.092   & 0.097 \\
NUDF ~\citep{sorensen2022nudf}                     &  0.587   &  1.885  &  0.524  & 0.272   & 0.416  & 0.115   & 0.158   & 0.115   & 0.095  & 0.100   &  0.410  &  1.064  & 0.226  & 0.107   & 0.186   & 0.115   &  0.135  &  0.106  &  0.092  & 0.097 \\
IOSNet ~\citep{khan2022implicit}                   &  0.999   &  1.352  &  0.742  & 0.572   & 0.679  & 0.132   & 0.142  & 0.132   & 0.105  & 0.122   & 0.458   & 0.932   & 0.685  & 0.202   & 0.526   & 0.124   & 0.125  & 0.125 &  0.098  & 0.117  \\  
CSR \citep{amiranashvili2024learning}             &   0.488   & 1.381   & 0.291   & 0.208    & 0.317   & 0.113   &  0.149   & 0.113    & 0.097   & 0.102    & 0.429  & 0.836   &  0.229  &  0.046  & 0.128    & 0.108     & 0.141   & 0.112  & 0.092   & 0.096   \\ \hlinew{0.50pt}
\textbf{Proposed}                                  &  \textbf{0.458}   &  \textbf{1.018}  & \textbf{0.177}   &  \textbf{0.174}  & \textbf{0.244}  & \textbf{0.101}   & \textbf{0.120}   & \textbf{0.094}  & \textbf{0.081}  & \textbf{0.094}   &  \textbf{0.283}  & \textbf{0.644}   & \textbf{0.175}  & \textbf{0.027}   & \textbf{0.113}   &  \textbf{0.093}  & \textbf{0.109}   & \textbf{0.093}   & \textbf{0.073}   & \textbf{0.077} \\
\hlinew{1.25pt}
\noalign{\vskip 1.5pt}
\multicolumn{21}{c}{\normalsize{\textbf{Voxel-wise Metric Space}}}\\[1.5pt]\hlinew{0.5pt}
\multirow{2}{*}{\textbf{Method}}& \multicolumn{5}{c}{\textbf{DSC (\%) $\uparrow$}} & \multicolumn{5}{c}{\textbf{NSD (\%) $\uparrow$}} & \multicolumn{5}{c}{\textbf{HD95 (mm) $\downarrow$}} & \multicolumn{5}{c}{\textbf{ASSD (mm) $\downarrow$}}  \\ \cmidrule(l){2-6} \cmidrule(l){7-11} \cmidrule(l){12-16} \cmidrule(l){17-21} 
                                                 &  RU   & RM   & RL   & LU   &  LL &  RU   & RM   & RL   & LU   &  LL    & RU   & RM   & RL   & LU   &  LL   & RU   & RM   & RL   & LU   &  LL  \\ \hline
Double-UNet ~\citep{jha2020doubleu}                &  92.62   &  83.98  &  95.03  &  95.76  & 95.36  &  78.27  &  68.51  & 85.59   & 91.13  & 89.73   & 16.37   & 21.26   & 15.55  & 11.32   & 15.35   & 2.44   & 3.92 & 1.88   & 1.22   &  1.57 \\
SCAN ~\citep{dai2018scan}                          &  92.57   &  84.41  & 95.14   &  95.37  &  95.08  & 81.19   &  \textbf{70.73}  & 86.42   & 89.79  & 89.52   & \textbf{13.68}   & 21.14   & 16.30  & 11.90   &  15.89  &  \textbf{2.05}  & 3.72  & 1.73   & 1.32   & 1.68  \\
ICTNet ~\citep{araujo2019deep}                     &  92.39  & 83.78   &  95.34  & 96.04   & 95.43  &  \textbf{82.14}  & 69.22   & 86.88   & 90.36  & 90.44   & 13.69   & 25.03   & 17.70  & 10.03   & 15.35   &  \textbf{2.05}  & 4.12 & 1.84   & 1.22   & 1.75 \\
Implicit U-Net ~\citep{marimont2022implicit}       &  91.98  & 75.38   & 94.34   & 93.78   &  92.48  &  78.69  & 62.61   & 84.32   &  86.58 &  86.31  & 19.99   & 20.49   & 15.87  &  15.95  &  19.85  & 3.22   & 4.85  & 2.03   & 2.21   & 2.44 \\
NUDF ~\citep{sorensen2022nudf}                     &  92.57  & \textbf{84.98}   & 86.25   &  95.77  &  96.05  &  80.72  &  69.92  & 89.75   & 91.67  & 92.06   & 24.04   & 36.70   &  14.77 & 13.73   & 9.56   & 2.77   & 5.14  &  1.40  & 1.37 & 1.18 \\
IOSNet ~\citep{khan2022implicit}                   &  90.30   & 72.89   & 91.56   & 91.57   & 89.00  & 73.65   & 59.13   & 78.29   & 78.24  & 77.86   & 21.98   & 21.83   & 19.95  & 22.64   & 21.55   & 4.10   & 5.30  & 2.89   & 3.34   & 4.00 \\
CSR \citep{amiranashvili2024learning}              &   92.08    &   83.94   & 95.33    &  95.66  & 95.70  &   81.18  &  70.17   &  89.61   &  92.91   & 92.24   & 14.49    & 21.02    & 11.81    &  9.31   & 9.85    & 2.43   &  3.79   & 1.52   & 1.07   & 1.42   \\ \hlinew{0.50pt}
\textbf{Proposed}                                  &  \textbf{93.60}   & 84.94   & \textbf{96.54}   & \textbf{96.37}   & \textbf{96.20}  & 81.70  & 70.38   & \textbf{90.44}   & \textbf{94.35}  & \textbf{93.48}   & 14.45   & \textbf{18.98} &  \textbf{10.01}  & \textbf{6.98} &  \textbf{8.34}  & 2.31   & \textbf{3.51}  & \textbf{1.01}   & \textbf{1.02}   & \textbf{1.03} \\
\hlinew{1.25pt}
\end{tabular}
\end{threeparttable}}
\end{table*}

\begin{figure}[t]
\centering
\includegraphics[width=0.6\linewidth]{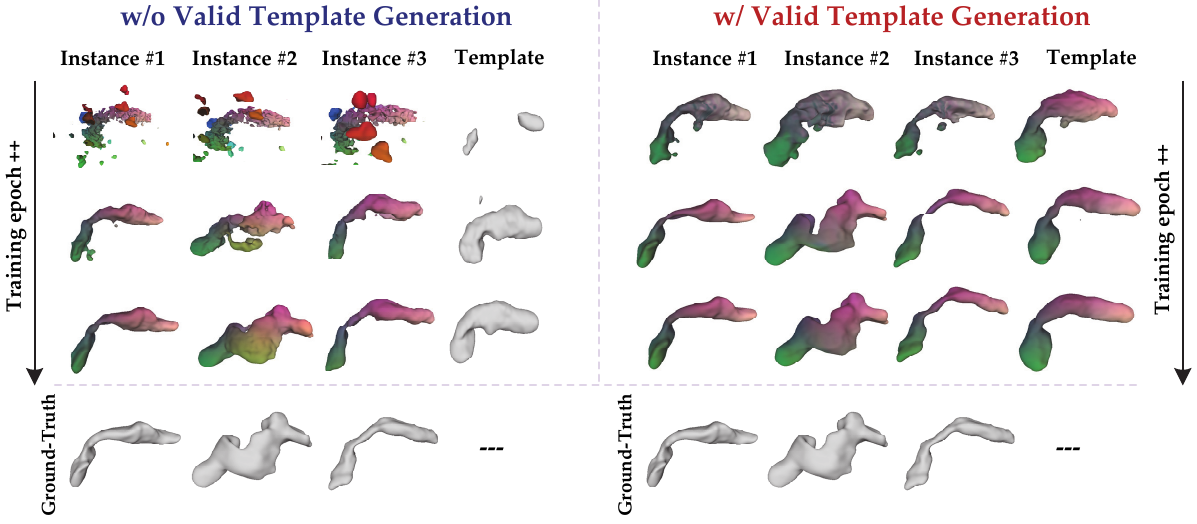}
\caption{The comparison of the implicit shape reconstruction and the template generation.}
\label{fig:valid_template_generation_comparation}
\end{figure}
\begin{figure}[t]
\centering
\includegraphics[width=0.5\linewidth]{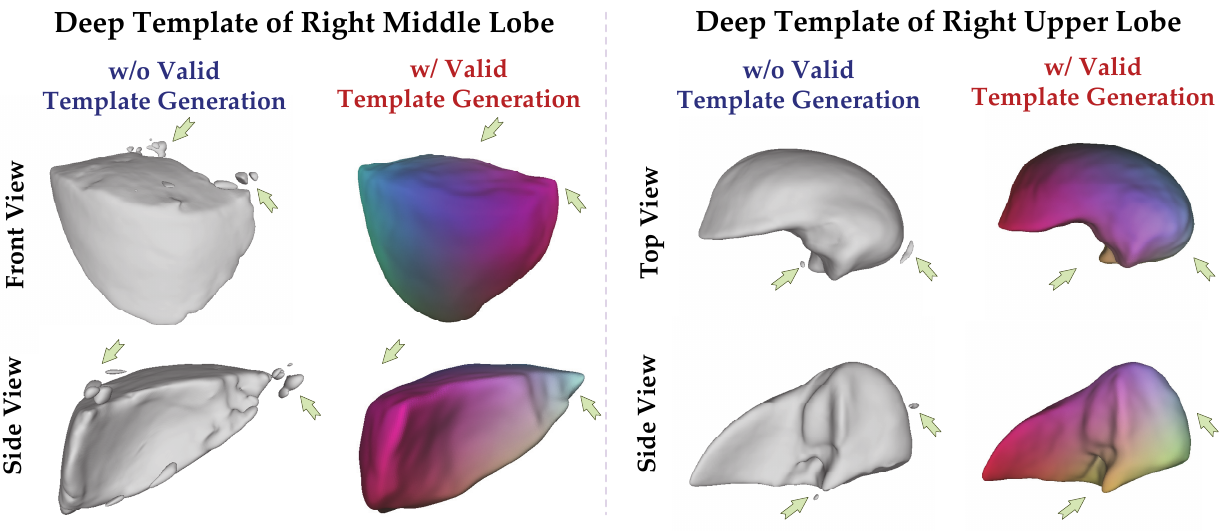}
\caption{Comparison of the validity of the generated implicit templates.}
\label{fig:valid_template_correctness_comparation}
\end{figure}

\tab{\ref{tab:continuous_wise_result_three_datasets}} and \tab{\ref{tab:multilobe_result}} report the quantitative results on the shape modeling of unseen CT data with regard to the measurements of CD and EMD. 
It is observed that our approach performed the best on shape modeling of different datasets, achieving average 0.225 CD / 0.085 EMD on Liver shape, 0.125 CD / 0.067 EMD on Pancreas shape, and 0.414 CD / 0.098 EMD on Lung Lobe shapes. Compared 
with the voxel-wise latent constrained methods~\citep{araujo2019deep,jha2020doubleu,dai2018scan}, they were limited in the discrete voxel representation, and the plausible structural prior knowledge is hard to acquire for implicit reconstruction. 
The qualitative result, seen in \fig{\ref{fig:reconstruction_error}}, also verified that these methods were hard to correct structural errors, e.g., Double UNet~\citep{jha2020doubleu} and SCAN~\citep{dai2018scan} generated anatomical 
error in Liver instances.As for the recent implicit-based segmentation methods~\citep{marimont2022implicit,sorensen2022nudf,khan2022implicit,amiranashvili2024learning}, ReshapeIT performed better than theirs because it 
not only breaks the limitation of the discrete voxel grid but also explores sharable structural prior knowledge for shape modeling. \fig{\ref{fig:reconstruction_error}} corroborated that the ReshapeIT achieved superior accuracy of shape modeling.

With regard to the implicit reconstruction procedure of multi lobes, \tab{\ref{tab:multilobe_result}} reported the quantitative results. The proposed method could decrease CD/EMD of each lobe to alleviate the misclassification problem. 
This can be credited to the independent template interaction and implicit reconstruction. As seen in \fig{\ref{fig:multi_lobes_refinement}}, we provided the example of the right upper lobe and right middle lobe. 
It was observed that misclassification happened in the initial results, however, the learned template for each lobe is reasonable for interaction. The template encodes each shape prior to the interaction, and the 
right upper lobe and right middle lobe had been rectified according to their independent templates. 
As demonstrated in \fig{\ref{fig:representative_reconstruction_refinement}} and \fig{\ref{fig:multi_lobes_refinement}}, the learned templates via deep INRs conform to the anatomical shape priors, 
hence the following reconstruction procedure shares more interpretability. 

As for the discrete space metrics when the reconstructed shape are inverted to the discrete grids.
Compared with the initial results generated by 3D UNet, the proposed implicit shape modeling refined the HD95 from 18.05 mm to 11.40 mm on the Liver dataset, from 4.91 mm to 4.13 mm on the Pancreas dataset, 
and from 14.97 mm to 12.04 mm on the Lung Lobe dataset. \tab{\ref{tab:discrete_wise_result_three_datasets}} demonstrated that the proposed method also performed better than other comparative methods, especially the surface distance based measurements HD95 and ASSD had been improved substantiality. \fig{\ref{fig:slice_comparison}} provided the slices view of 
the comparison between the initial results and implicit reconstruction results. It can be observed that the under/over segment problem had been alleviated with the integration of template anatomical prior. In addition, 
the implicit reconstruction was performed in the continuous space, hence, the response near the boundary became more consistent, manifested as smoother.


\makeatletter
\def\hlinew#1{%
\noalign{\ifnum0=1}\fi\hrule \@height #1 \futurelet
\reserved@a\@xhline}
\makeatother
\begin{table*}[t]
\renewcommand\arraystretch{1.0}
\centering
\caption{Ablation study of the instance latent code $\bm{\alpha}$ and hyper latent code $\bm{\beta}$ on Liver, Pancreas, and Lung Lobe datasets. Mean CD and EMD are reported.}\label{tab:Ablation_latent_code}
\scalebox{0.5}{
\begin{threeparttable}
\begin{tabular}{cccccccc}
\hlinew{1.25pt}
&            & \multicolumn{2}{|c|}{\textbf{Liver}} & \multicolumn{2}{c|}{\textbf{Pancreas}} & \multicolumn{2}{c}{\textbf{Lung Lobe}} \\ 
Instance code $\bm{\alpha}$ & Hyper code $\bm{\beta}$ &  \multicolumn{1}{|c}{CD$\downarrow$}    & \multicolumn{1}{c|}{EMD$\downarrow$}             & CD     $\downarrow$          & \multicolumn{1}{c|}{EMD $\downarrow$    }            & CD     $\downarrow$           & EMD    $\downarrow$           \\ \hline

\red{\XSolidBrush}&\red{\XSolidBrush}  & \multicolumn{1}{|c}{0.340 }& \multicolumn{1}{c|}{0.101} &   0.184              &    \multicolumn{1}{c|}{0.087}           &   0.617                &    0.110               \\
\green{\Checkmark}&\red{\XSolidBrush}  & \multicolumn{1}{|c}{0.259} & \multicolumn{1}{c|}{0.089} &   0.130              &    \multicolumn{1}{c|}{0.071}             &      0.483             &   0.104               \\
\red{\XSolidBrush}&\green{\Checkmark}  & \multicolumn{1}{|c}{0.271 }& \multicolumn{1}{c|}{0.093} &   0.136             &   \multicolumn{1}{c|}{0.074}             &       0.507            &   0.105                 \\
\green{\Checkmark}&\green{\Checkmark}  & \multicolumn{1}{|c}{\textbf{0.225}} & \multicolumn{1}{c|}{\textbf{0.085}} &   \textbf{0.125}   &    \multicolumn{1}{c|}{\textbf{0.067}}     &   \textbf{0.414}  &  \textbf{0.098}     \\
\hlinew{1.25pt}
\end{tabular}
\end{threeparttable}}
\end{table*}

\subsection{Analysis of the Valid Template Generation}
\fig{\ref{fig:valid_template_generation_comparation}} and \fig{\ref{fig:valid_template_correctness_comparation}} 
illustrate the effectiveness of the instance-template constraint used in the ReshapeIT to generate valid templates.
It can be observed that optimization procedure shares difference when with or without the valid template generation. 
Without the valid template generation, the implicit network converges to fit the shapes of the instances, while neglecting 
the correctness of the latent template shape. Fortunately, the template could be more accurately generated upon strengthening 
the correspondence between the instance and the template, as demonstrated in \fig{\ref{fig:valid_template_generation_comparation}}. 
In addition, \fig{\ref{fig:valid_template_correctness_comparation}} displays that some floating artifacts can be observed in the 
template generation of the right middle lobe and the right upper lobe when not using the instance-template constraint. 
These floating artifacts can be addressed via constraining the implicit template shape to be valid as the same for instances.

\makeatletter
\def\hlinew#1{%
\noalign{\ifnum0=1}\fi\hrule \@height #1 \futurelet
\reserved@a\@xhline}
\makeatother
\begin{table*}[t]
\renewcommand\arraystretch{1.0}
\centering
\caption{Plugin usage of the proposed ReShapeIT. Quantitative continuous metric space results on Liver, Pancreas, and Lung Lobe datasets.}\label{tab:plugin_continuous_three_datasets}
\scalebox{0.5}{
\begin{threeparttable}
\begin{tabular}{lcccc}
\hlinew{1.25pt}
\noalign{\vskip 1.5pt}
\multicolumn{5}{c}{\normalsize{\textbf{Liver Dataset}}}\\[1.5pt]\hlinew{0.5pt}
\textbf{Method} & \textbf{CD (mean. $\pm$ std.) $\downarrow$} & \textbf{EMD (mean. $\pm$ std.) $\downarrow$} & \textbf{CD (median) $\downarrow$} & \textbf{EMD (median) $\downarrow$}  \\ \hline
UNETR ~ \citep{hatamizadeh2022unetr}  & 0.649 $\pm$ 0.387 & 0.144 $\pm$ 0.034 & 0.258 & 0.117 \\
UNETR  w/ ReShapeIT & \textbf{0.577 $\pm$ 0.294} & \textbf{0.121 $\pm$ 0.028} & \textbf{0.457} & \textbf{0.107} \\\hlinew{0.25pt}
nnUNet ~\citep{isensee2021nnu} & 0.128 $\pm$ 0.118 & 0.089 $\pm$ 0.024 & 0.078 & 0.089 \\
nnUNet w/ ReShapeIT & \textbf{0.077 $\pm$ 0.067}  & \textbf{0.056 $\pm$ 0.015}  & \textbf{0.055}  & \textbf{0.052}  \\ 
\hlinew{1.25pt}
\noalign{\vskip 1.5pt}
\multicolumn{5}{c}{\normalsize{\textbf{Pancreas Dataset}}}\\[1.5pt]\hlinew{0.5pt}
\textbf{Method} & \textbf{CD (mean. $\pm$ std.) $\downarrow$} & \textbf{EMD (mean. $\pm$ std.) $\downarrow$} & \textbf{CD (median) $\downarrow$} & \textbf{EMD (median) $\downarrow$}  \\ \hline
UNETR ~ \citep{hatamizadeh2022unetr}  & 1.918 $\pm$ 2.557 & 0.136 $\pm$ 0.063 & 1.056 & 0.121 \\
UNETR  w/ ReShapeIT    & \textbf{1.442 $\pm$ 1.545} & \textbf{0.107 $\pm$ 0.041} & \textbf{0.887} & \textbf{0.093} \\ \hlinew{0.25pt}
nnUNet ~\citep{isensee2021nnu} & 0.128 $\pm$ 0.118 & 0.089 $\pm$ 0.024 & 0.078 & 0.089\\
nnUNet w/ ReShapeIT & \textbf{0.075 $\pm$ 0.066}  & \textbf{0.057 $\pm$ 0.014}  & \textbf{0.054}  & \textbf{0.051}  \\
\hlinew{1.25pt}
\noalign{\vskip 1.5pt}
\multicolumn{5}{c}{\normalsize{\textbf{Lung Lobe Dataset}}}\\[1.5pt]\hlinew{0.5pt}
\textbf{Method} & \textbf{CD (mean. $\pm$ std.) $\downarrow$} & \textbf{EMD (mean. $\pm$ std.) $\downarrow$} & \textbf{CD (median) $\downarrow$} & \textbf{EMD (median) $\downarrow$}  \\ \hline
UNETR ~ \citep{hatamizadeh2022unetr}  & 0.633 $\pm$ 0.640 & 0.113 $\pm$ 0.024 & 0.422 & 0.117 \\
UNETR  w/ ReShapeIT   & \textbf{0.602 $\pm$ 0.617} & \textbf{0.107 $\pm$ 0.022} & \textbf{0.390}  & \textbf{0.112}  \\ \hlinew{0.25pt}
nnUNet ~\citep{isensee2021nnu} & 0.361 $\pm$ 0.520 & 0.105 $\pm$ 0.023 & 0.163  & 0.101 \\
nnUNet w/ ReShapeIT & \textbf{0.186 $\pm$ 0.235}   & \textbf{0.088 $\pm$ 0.015}   &  \textbf{0.121}   & \textbf{0.084}  \\
\hlinew{1.25pt}
\end{tabular}
\end{threeparttable}}
\end{table*}

\makeatletter
\def\hlinew#1{%
\noalign{\ifnum0=1}\fi\hrule \@height #1 \futurelet
\reserved@a\@xhline}
\makeatother
\begin{table*}[htbp]
\renewcommand\arraystretch{1.0}
\centering
\caption{Plugin usage of the proposed method. Quantitative discrete metric space results on Liver, Pancreas, and Lung Lobe datasets.}
\label{tab:plugin_discrete_three_datasets}
\scalebox{0.5}{
\begin{threeparttable}
\begin{tabular}{lcccc}
\hlinew{1.25pt}
\noalign{\vskip 1.5pt}
\multicolumn{5}{c}{\normalsize{\textbf{Liver Dataset}}}\\[1.5pt]\hlinew{0.5pt}
\textbf{Method} & \textbf{DSC (\%) $\uparrow$} & \textbf{NSD (\%) $\uparrow$} & \textbf{HD95 (mm) $\downarrow$} & \textbf{ASSD (mm) $\downarrow$}  \\ \hline
UNETR ~ \citep{hatamizadeh2022unetr}  & 90.72 $\pm$ 6.42 & 78.13 $\pm$ 14.65 & 25.88 $\pm$ 23.46 & 3.14 $\pm$ 3.07 \\
UNETR  w/ ReShapeIT  & \textbf{91.45 $\pm$ 6.04} & \textbf{80.21 $\pm$ 10.51} & \textbf{23.65 $\pm$ 21.14}  & \textbf{2.84 $\pm$ 2.68} \\ \hlinew{0.25pt}
nnUNet ~\citep{isensee2021nnu} & 95.23 $\pm$ 4.25 & 91.08 $\pm$ 8.54 & 9.88 $\pm$ 9.01 & 1.37 $\pm$ 1.13\\
nnUNet w/ ReShapeIT & \textbf{95.51 $\pm$ 3.05} & \textbf{93.46 $\pm$ 7.15}  & \textbf{8.50 $\pm$ 6.04} & \textbf{1.25 $\pm$ 0.94}\\
\hlinew{1.25pt}
\noalign{\vskip 1.5pt}
\multicolumn{5}{c}{\normalsize{\textbf{Pancreas Dataset}}}\\[1.5pt]\hlinew{0.5pt}
\textbf{Method} & \textbf{DSC (\%) $\uparrow$} & \textbf{NSD (\%) $\uparrow$} & \textbf{HD95 (mm) $\downarrow$} & \textbf{ASSD (mm) $\downarrow$}  \\ \hline
UNETR ~ \citep{hatamizadeh2022unetr}  & 78.98 $\pm$ 8.24 & 87.10 $\pm$ 7.38 & 20.68 $\pm$ 22.67 & 2.69 $\pm$ 2.31 \\
UNETR  w/ ReShapeIT   & \textbf{79.84 $\pm$ 7.95} & \textbf{88.05 $\pm$ 6.98} & \textbf{18.56 $\pm$ 20.95} & \textbf{1.94 $\pm$ 1.88}  \\ \hlinew{0.25pt}
nnUNet ~\citep{isensee2021nnu} & 85.78 $\pm$ 5.10 & 94.41 $\pm$ 4.46 & 3.90 $\pm$ 2.28 & 0.90 $\pm$ 0.36\\
nnUNet w/ ReShapeIT & \textbf{86.45 $\pm$ 5.46} & \textbf{94.65 $\pm$ 6.19}  & \textbf{3.81 $\pm$ 2.71} & \textbf{0.86 $\pm$ 0.40}\\
\hlinew{1.25pt}
\noalign{\vskip 1.5pt}
\multicolumn{5}{c}{\normalsize{\textbf{Lung Lobe Dataset}}}\\[1.5pt]\hlinew{0.5pt}
\textbf{Method} & \textbf{DSC (\%) $\uparrow$} & \textbf{NSD (\%) $\uparrow$} & \textbf{HD95 (mm) $\downarrow$} & \textbf{ASSD (mm) $\downarrow$}  \\ \hline 
UNETR ~ \citep{hatamizadeh2022unetr}  & 92.74 $\pm$ 5.00 & 83.73 $\pm$ 10.18 & 14.37 $\pm$ 10.32 & 2.04 $\pm$ 1.46 \\
UNETR  w/ ReShapeIT & \textbf{93.14 $\pm$ 4.82} & \textbf{84.04 $\pm$ 9.85} & \textbf{14.08 $\pm$ 9.94} & \textbf{1.95 $\pm$ 1.25} \\ \hlinew{0.25pt}
nnUNet  ~\citep{isensee2021nnu} & \textbf{95.78 $\pm$ 3.43} & 90.82 $\pm$ 7.28 & 7.98 $\pm$ 9.33  & 1.07 $\pm$ 0.84  \\
nnUNet w/ ReShapeIT & 95.73  $\pm$ 2.04 & \textbf{91.00 $\pm$ 4.53}  & \textbf{7.66 $\pm$ 4.26}  & \textbf{0.96 $\pm$ 0.52} \\ 
\hlinew{1.25pt}
\end{tabular}
\end{threeparttable}}
\end{table*}

\section{Discussion}
\subsection{Ablation Study}
\noindent\textbf{Effectiveness of $\bm{\alpha}$ and $\bm{\beta}$:} To valid the effect of components, $\bm{\alpha}$, and $\bm{\beta}$, we conducted the ablation study in \tab{\ref{tab:Ablation_latent_code}}. 
Without instance code $\bm{\alpha}$, the CD/EMD suffered a performance decline. CD/EMD got worse results to 0.271/0.093 in the Liver dataset, 0.136/0.074 in the Pancreas dataset, and 0.507/0.105 in the Lobe dataset. It demonstrated 
the necessity to assign a unique latent code for each instance. Similarly, Without Hyper code $\bm{\beta}$, the CD/EMD also encountered the degradation in CD/EMD to 0.259/0.089 in the Liver dataset, 
0.130/0.071 in the Pancreas dataset, and 0.483/104 in the Lobe dataset. The $\bm{\beta}$ that forms the hyper-network is helpful to encourage the weights of the deformation field to be instance-specific and better fuse 
the instance latent codes and corresponding coordinates. Moreover, the effectiveness of the regularization constraints can be found in the Supplementary Material.


\noindent\textbf{Plug \& Play Usage:} We also demonstrated that the proposed implicit shape modeling of volumetric data shares the plug-and-play property. We embedded the implicit shape modeling to refine the 
initial results generated by the nnUNet~\citep{isensee2021nnu} and UNETR \citep{hatamizadeh2022unetr}. 
nnUNet is a strong tool for medical image segmentation and UNETR is the representative method that uses the transformer as the backbone. 
However, it still lacks the shape prior knowledge during the optimization process. 
\tab{\ref{tab:plugin_continuous_three_datasets}} and \tab{\ref{tab:plugin_discrete_three_datasets}} revealed that the proposed ReShapeIT can consistently further improve the initial results of nnUNet and UNETR.

\subsection{More Challenges and Future}
One of the promising applications of the proposed method is to refine the imperfect annotation of volumetric data, which can substantially relieve the burden of precise labeling. It is feasible because the class-wise template shape prior has been learned in the continuous space across all collected instances. Further, this template shape prior could bring guidance into the imperfect annotation of unseen data to perform the implicit reconstruction. 
In addition, shape registration can be performed based on the implicit neural fields. Each instance is wrapped to the template space via an implicit deformation field. Hence, the dense correspondences between instances could be built through the connection of the implicit template. 
The implicit shape modeling with template registration provides the registered 3D shapes, which can be further transformed into virtual anatomical models. These virtual models are beneficial to the AR-based anatomy education~\citep{bolek2021effectiveness}.

Our work focused on the implicit shape modeling of the medical organs that share common structures within the same category. However, affected by the diseases or surgeries, the anatomical structure of organs can be changed 
for patients, for example, the hepatic segmentectomy would change the original liver shape. The current learned implicit template cannot be well generalized to specific cases with non-existent partial structures. 
One potential future solution is to extend the implicit representation to jointly learn a global shape prior with local anatomical-varying structures conditioned by the image intensity distribution. In addition, 
the current approach generates independent implicit templates for multi-class medical shapes. To model the multi-class shapes in a unified implicit neural network is challenging. For example, the lobes interact 
with each other closely, leading to the accurate SDF values are difficult to acquire since the fissure interface is hardly visible from any camera view. Hence, to compensate for SDF values, 
other supervision signals should be further explored to characterize the representation of the implicit field.

\section{Conclusion}
In this paper, we propose the ReShapeIT network, which models anatomical structures in continuous space rather than discrete voxel grids. 
ReShapeIT represents an anatomical structure with an implicit template field shared within the same category, along with a deformation field.
The constraint of the correspondence between the instance shape and the template shape is strengthened to ensure the generation of valid templates. 
The valid template shape can then be used for implicit generalization via the TIM module.
Experimental results on three datasets demonstrated the superiority of our approach in anatomical structure reconstruction compared with other state-of-the-art methods.

\section{Acknowledgement}
This work was supported in part by National Key R\&D Program of China (Grant Number: 2022ZD0212400), Natural Science Foundation of China (Grandt Number: 62373243) and the Science and Technology Commission of Shanghai Municipality (Grant Number: 20DZ2220400), Shanghai Municipal Science and Technology Major Project (No.2021SHZDZX0102).

\newpage

\section{Supplementary Materials}

\subsection{Evaluation Metrics}
Chamfer Distance (CD) \citep{fan2017point} 
and Earth Mover's Distance (EMD) \citep{rubner1998metric} are two appropriate metrics to measure shapes represented by meshes. These two metrics are calculated on two point sets $S_{1}$ and $S_{2}$ 
that are sampled from the surfaces of refined shape and ground-truth respectively. The CD can be defined as:
\begingroup
\small
\begin{align}
\begin{aligned}
\mathrm{CD}  &\overset{\text{def}}{=} d_{CD}(S_{1},S_{2})  \\ & = \sum_{x \in S_{1}} \min_{y \in S_{2}}\left \| x - y \right \|_{2}^{2} + \sum_{y \in S_{2}} \min_{x \in S_{1}}\left \| y - x \right \|_{2}^{2}. \label{eq:CD}  
\end{aligned}
\end{align}
\endgroup
In all experiments, we used 30K points for both $S_{1}$ and $S_{2}$, and reported CD by the normalized result: $\frac{d_{CD}(S_{1},S_{2})}{30,000}$. With regard to the EMD~\citep{rubner1998metric}, which forms 
a bijection: $\phi:S_{1} \to S_{2}$. The EMD is defined based on the optimal bijection:
\begingroup
\small
\begin{align}
\mathrm{EMD} \overset{\text{def}}{=} d_{EMD}(S_{1},S_{2}) =   \min_{\phi: S_{1} \to S_{2}}\sum_{x \in S_{1}}\left \| x - \phi(x) \right \|_{2}^{2}. \label{eq:EMD} 
\end{align}
\endgroup
We adopted 1K points to calculate EMD in all experiments.

\begin{figure}[t]
\centering
\includegraphics[width=0.8\linewidth]{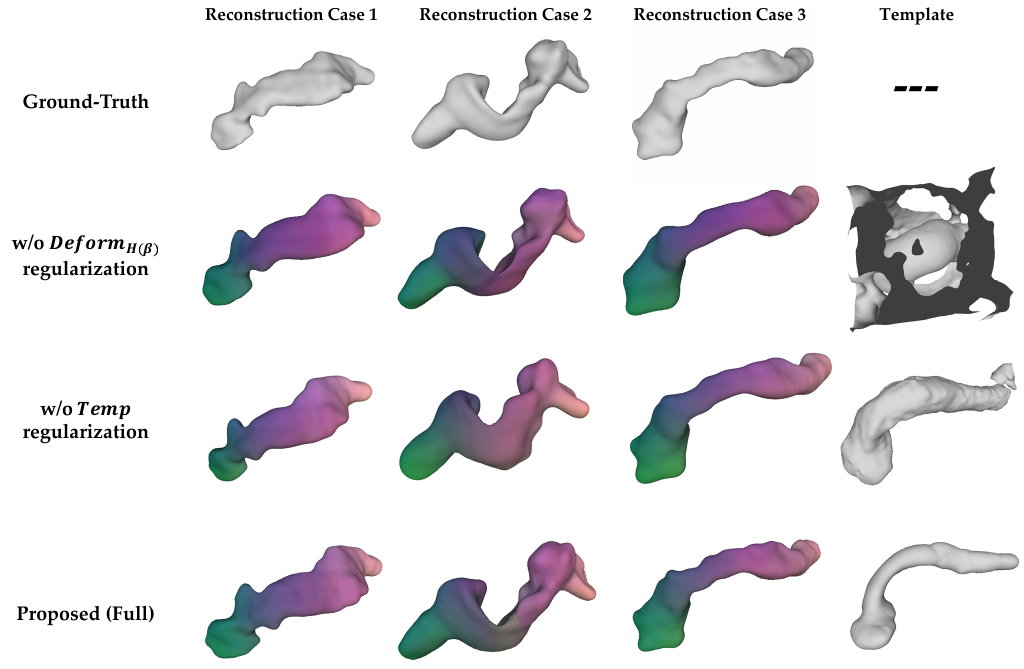}
\caption{Effect of the regularization constraints of the $\mathit{Deform}_{\mathcal{H}(\bm{\beta})}$ and $\mathit{Temp}$.}
\label{fig:regularization_deform_temp}
\end{figure}

\subsection{Effectiveness of Regularization Constraints}
To investigate the role of the regularization constraints, we performed the ablation study of $\mathit{Deform}_{\mathcal{H}(\bm{\beta})}$ and $\mathit{Temp}$. 
\fig{\ref{fig:regularization_deform_temp}} visualized the experimental results. For one thing, without the regularization of $\mathit{Deform}_{\mathcal{H}(\bm{\beta})}$, i.e., $\mathcal{L}_{vec}$ and $\mathcal{L}_{dis}$, the instance-specific 
shape modeling achieved satisfactory results. However, the loose restriction on $\mathit{Deform}_{\mathcal{H}(\bm{\beta})}$ leads to the invalid template shape, as illustrated in the second row of \fig{\ref{fig:regularization_deform_temp}}, 
which is detrimental to the subsequent template interaction and implicit reconstruction process. For another, the lack of regularization of $\mathit{Temp}$, i.e., $\mathcal{L}_{tpn}$, caused less accurate shape modeling of 
the instances and the sharable template structure cannot be extracted with high fidelity across the instances within the same category. In summary, $\mathit{Deform}_{\mathcal{H}(\bm{\beta})}$ and $\mathit{Temp}$ are 
two indispensable regularization constraints to guarantee accurate implicit shape modeling with template generation.

\bibliographystyle{elsarticle-num.bst}
\bibliography{ref}
\end{document}